\documentclass[journal]{IEEEtran}
\usepackage{xcolor,soul,framed} 
\colorlet{shadecolor}{yellow}
\usepackage[pdftex]{graphicx}
\graphicspath{{../pdf/}{../jpeg/}}
\DeclareGraphicsExtensions{.pdf,.jpeg,.png}
\usepackage[cmex10]{amsmath}
\usepackage{array}
\usepackage{amssymb}
\usepackage{mdwmath}

\usepackage{mdwtab}
\usepackage{eqparbox}
\usepackage{url}
\usepackage{enumerate}
\usepackage{enumitem}
\usepackage{amsfonts}
\usepackage[english]{babel}
\usepackage[utf8]{inputenc}
\usepackage[noend]{algpseudocode}
\usepackage{mathtools}
\usepackage{nomencl}
\usepackage{blindtext}
\usepackage{amsmath}
\usepackage[colorinlistoftodos]{todonotes}
\usepackage{color}
\usepackage{color,soul}
\usepackage{xcolor}
\usepackage{multirow}
\usepackage{graphicx}
\usepackage{epstopdf}
\usepackage[T1]{fontenc}
\usepackage{textcase}
\usepackage{array}
\usepackage{mdwmath}
\usepackage{mdwtab}
\usepackage{eqparbox}
\usepackage{enumerate}
\usepackage{bm}
\usepackage{amsfonts}
\usepackage[ruled,lined,linesnumbered]{algorithm2e}
\usepackage{booktabs}
\usepackage{comment}
\usepackage{amsmath}
\usepackage{mathtools}
\usepackage{caption}
\usepackage{subcaption}
\usepackage{tabularx}
\usepackage{hyperref}
\hypersetup{
    colorlinks=true,
    linkcolor=blue,
    filecolor=magenta,      
    urlcolor=cyan,
}

\usepackage{caption}
\usepackage{mathrsfs}
\usepackage{booktabs}
\usepackage{siunitx}
\usepackage{tabularx}

\usepackage{cite}
\usepackage{url}
\usepackage{setspace}
\usepackage{amsthm}
\usepackage{nomencl}
\usepackage{footnote}
\usepackage{threeparttable}
\usepackage{xcolor}

\makenomenclature
\usepackage{etoolbox}
\renewcommand\nomgroup[1]{%
  \item[\bfseries
  \ifstrequal{#1}{A}{Abbreviations}{%
  \ifstrequal{#1}{P}{Power System Operation}{%
  \ifstrequal{#1}{M}{Machine Unlearning}{%
  \ifstrequal{#1}{S}{Symbols}{}}}}%
]}

\newtheorem{proposition}{Proposition}

\usepackage{xcolor,cite,etoolbox}
\makeatletter 
\pretocmd\@bibitem{\color{black}\csname keycolor#1\endcsname}{}{\fail}

\usepackage{etoolbox}
\makeatletter
\patchcmd{\@makecaption}
  {\scshape}
  {}
  {}
  {}
\makeatother

\title{Task-Aware Machine Unlearning and Its Application in Load Forecasting}

\author{Wangkun Xu,~\IEEEmembership{Student Member,~IEEE},
    and Fei Teng,~\IEEEmembership{Senior Member,~IEEE}\\
    \thanks{
    \\
    Wangkun Xu and Fei Teng are with the Department of Electrical and Electronic Engineering, Imperial College London, UK 
    
(\textit{Corresponding author: Fei Teng}, email: f.teng@imperial.ac.uk). 
    }
}

\begin{document}
\setlength{\textfloatsep}{1pt}
\renewcommand{\baselinestretch}{1}
\markboth{This paper has been accepted by IEEE trans on power systems. copyright of the paper is reserved by ieee}%
{Shell \MakeLowercase{\textit{et al.}}: Bare Demo of IEEEtran.cls for IEEE Journals}
\maketitle

\begin{abstract}

Data privacy and security have become a non-negligible factor in load forecasting. Previous researches mainly focus on training stage enhancement. However, once the model is trained and deployed, it may need to `forget' (i.e., remove the impact of) part of training data if the these data are found to be malicious or as requested by the data owner. This paper introduces the concept of machine unlearning which is specifically designed to remove the influence of part of the dataset on an already trained forecaster. However, direct unlearning inevitably degrades the model generalization ability. To balance between unlearning completeness and model performance, a performance-aware algorithm is proposed by evaluating the sensitivity of local model parameter change using influence function and sample re-weighting. Furthermore, we observe that the statistical criterion such as mean squared error, cannot fully reflect the operation cost of the downstream tasks in power system. Therefore, a task-aware machine unlearning is proposed whose objective is a trilevel optimization with dispatch and redispatch problems considered. We theoretically prove the existence of the gradient of such an objective, which is key to re-weighting the remaining samples. We tested the unlearning algorithms on linear, CNN, and MLP-Mixer based load forecasters with a realistic load dataset. The simulation demonstrates the balance between unlearning completeness and operational cost.
All codes can be found at \url{https://github.com/xuwkk/task_aware_machine_unlearning}.

\end{abstract}


\mbox{}
\setlength{\nomlabelwidth}{2cm}
\nomenclature[A]{PA/TA-MU}{Performance/Task Aware Machine Unlearning}
\nomenclature[A]{NN}{Neural Network}
\nomenclature[A]{MLP}{Multi Layer Perceptron}
\nomenclature[A]{CNN}{Convolutional Neural Network}
\nomenclature[A]{MSE}{Mean Squared Error}
\nomenclature[A]{MAPE}{Mean Absolute Percentage Error}
\nomenclature[A]{SO}{System Operator}

\nomenclature[M]{$\mathcal{D}$, $\mathcal{D}_{\text{test}}$, $\mathcal{D}_{\text{unlearn}}$, $\mathcal{D}_{\text{remain}}$}{The training, test, unlearn, and remain dataset}
\nomenclature[M]{$\bm{f}(\cdot;\bm{\theta})$}{Load forecast model}
\nomenclature[M]{$\ell^i(\cdot)$, $\ell^i_{\text{test}}(\cdot)$}{The training and test loss/criterion of the $i$-th sample}
\nomenclature[M]{$\mathcal{L}(\cdot)$, $\mathcal{L}_{\text{test}}(\cdot)$}{The training and test cost function}
\nomenclature[M]{$\bm{y}^i$, $\hat{\bm{y}}^i$}{The ground-truth and forecast load of the $i$-th sample}
\nomenclature[M]{$N$, $N_{\text{test}}$}{The size of training and test dataset}
\nomenclature[M]{$\epsilon^i$}{The weight on the $i$-th sample}
\nomenclature[M]{$\lambda_1$, $\lambda_\infty$}{The 1-norm and inf-norm limits on the sample weights variation}
\nomenclature[M]{$\bm{\theta}^\star$, $\bm{\theta}_{\text{mod}}^\star$, $\bm{\theta}_{\text{remain}}^\star$}{Optimal Model Parameters}

\nomenclature[P]{$\bm{P}_g$, $\bm{P}_{ls}$, $\bm{P}_{gs}$}{The vector of generator set-points, load sheddings, and energy storages}
\nomenclature[P]{$\bm{Q}_g$, $\bm{c}_g$}{Second and first order coefficient of the generator cost}
\nomenclature[P]{$c_{ls2}$, $c_{ls}$}{Second and first order coefficient of load shedding}
\nomenclature[P]{$c_{gs2}$, $c_{gs}$}{Second and first order coefficient of energy storage}

\printnomenclature

\section{Introduction}



\subsection{Data Privacy and Security in Load Forecasting}

\IEEEPARstart{A}{ccurate} load forecasting is essential for the security and economic operation of the power system. The deterministic \cite{xie2015long} and probabilistic \cite{hong2016probabilistic} methods are two main categories. Recently, machine and deep learning algorithms have been widely applied to better retrieve spatial and temporal information, which certainly benefit the progression of load forecasting \cite{hong2020energy}. To fulfill the training purpose, large amount of data is collected from individuals, which challenges the integrity of data ownership and security. 

In power system, the system operator (SO) collects and transfers individual data for various operational purposes. However, this arrangement has raised privacy concerns, as individual load data are sensitive and can be targeted to retrace personal identity and behavior \cite{ebeid2017deducing}. From the perspective of data security, data collected from unsecured sources are prone to errors and adversaries. For example, the authors of \cite{luo2018benchmarking} benchmark how poor training data could degrade forecast accuracy by introducing random noise. Furthermore, data poisoning attack is specifically designed to contaminate the training dataset to prevent the load forecaster from being accurate at the test stage \cite{liang2019poisoning}.

Most of the existing work designs a preventive training algorithm to address concerns about data privacy and security. For example, federated learning is studied, in which each training participant only shares the trained parameters with the central server \cite{wang2022personalized}. In \cite{dong2022short}, a fully distributed training framework has been proposed in which each participant only shares the parameters with his neighbors. Differential privacy is another privacy-preserving technique used for load forecasting to avoid identifying the individual \cite{soykan2019differentially}. To combat poisoning attack, federated learning enhanced with differential privacy is developed in \cite{fernandez2022privacy}. By weight-clipping and adding noise to the central parameter update, the global model can be resistant to inference attacks to some extent. In addition, gradient quantization is applied, where each participant only uploads the sign of the local gradient \cite{husnoo2023secure}. 

\subsection{Machine Unlearning}

However, training stage prevention is not sufficient when the post-action of removing the impact of those data from the trained forecaster is needed. From privacy concerns, in addition to the right to share the data, many national and regional regulations have certified the consumers' \textit{`right to forget'} \cite{mantelero2013eu}, such as the European Union's General Data Protection Regulation (GDPR) and the recent US's California Consumer Privacy Act (CCPA). That is, consumers are eligible to request to destroy their personal records at any stage of the service, including the encoded information in the trained model\cite{shaik2023exploring}. Meanwhile, the SO may not be aware of the data defect until the model has been trained and deployed. Obviously, a straightforward approach to removing the impact of part of the dataset is to retrain the model from scratch on the remaining data. However, retraining can be computationally expensive and sometimes infeasible due to the lack of the original dataset.

In this context, machine unlearning (MU) has been introduced in machine learning especially the computer vision community to study the problem of removing a subset of training data, the forget or unlearn dataset, from the trained model. It has recently been extended to other practical fields, such as removing bias in language models \cite{yu2023unlearning}, unlearning personal information on the Internet of Things \cite{zeng2023towards} and digital twin mobile networks \cite{xia2023fedme}, as well as removing malicious samples in wireless communication beam selection problems \cite{zhang2023poison}.

Originating from \cite{cao2015towards} for statistical query learning, MU can be broadly classified into exact and approximate unlearning. Exact unlearnings are developed for specific algorithms, such as k-means \cite{ginart2019making} and modified random forest \cite{brophy2021machine}. The gradient and the Hessian matrix of the training objective are useful to approximate the influence of samples on the parameters of the trained model. Therefore,  the Fisher information \cite{golatkar2020eternal, peste2021ssse} and the influence function \cite{fu2021bayesian, golatkar2021mixed, guo2019certified} are adopted to unlearn the influence of the forget dataset from the trained model. Motivated by differential privacy, \cite{guo2019certified} certifies the exactness of data removal in linear classifiers. However, these methods are difficult to generalize to the neural network (NN) \cite{bae2022if} with guaranteed unlearning performance. To overcome the problem, a mixed-privacy forgetting is proposed to only unlearn on a linear regression model around the trained NN \cite{golatkar2021mixed, guo2019certified}. Projected gradient unlearning is proposed in \cite{tuan2024learn}. The gradient orthogonal to the column space of gradients of the remaining dataset is adopted to incrementally unlearn the forget dataset without catastrophically forgetting the remaining dataset.

Another line of research assumes that the model is trained with an oracle, that is, by taking into account the future unlearning requirement during training. For example, amnesiac training tracks the contribution of each training batch. When data in a batch is requested to be removed, the batch contribution can simply be subtracted \cite{graves2021amnesiac}. Alternatively, an exact but efficient retraining algorithm is proposed in \cite{bourtoule2021machine} in which ensemble models are trained on disjoint subsets of data. Therefore, only the model trained on the unlearned dataset needs to be re-trained. However, when the forget dataset spreads over multiple models, this method becomes less efficient. Finally, the theorem and application of machine unlearning are continually studied and more information can be found in the recent review \cite{nguyen2022survey}.

\subsection{Research Gaps}

\subsubsection{Unlearning Completeness vs Model Performance}

Although retraining is usually not a viable option, it is broadly agreed that the \textit{golden rule for unlearning} is to minimize the distance between unlearnt and retrained models \cite{nguyen2022survey}. In addition, the unlearning algorithm is \textit{complete} if the unlearnt model is identical to the model re-trained on the remaining dataset. However, we argue that complete unlearning may not be suitable for power system applications. Referring to Table \ref{tab:unlearning}, when the privacy is mainly concerned (\textit{privacy-driven MU}), although complete unlearning can certainly remove the influence of the forget dataset, it can inevitably degrade the performance of the trained model so that the interest of the remaining customers is harmed \cite{shaik2023exploring}. For the \textit{security-driven MU}, the malicious data can still contain useful information. However, the \textit{complete unlearning} not only removes the adverse influence of the forget dataset, but also the useful one.

\begin{table}[h]
    \centering
    \footnotesize
    \caption{The Purposes of Unlearning}
    \begin{tabularx}{\linewidth}{l|X|X}
       \textbf{MU Purpose}  & \textbf{Description} & \textbf{Target}  \\\hline\hline
       Privacy-driven  & Some training data contains sensitive information and is asked to remove by the customer. & The \textbf{main} target is to unlearn the model as if it is originally trained without the forget data. \\\hline
       Security-driven & Some training data is malicious or biased, whose influence should be removed from the trained model by the SO. & The \textbf{main} target is to remove the malicious information from the model while keeping the useful information.
    \end{tabularx}
    \label{tab:unlearning}
\end{table}

Therefore, under both privacy and security concerns, machine unlearning is to eliminate the influence of the data from the load forecaster while considering the possible influence on the model performance. We model this dilemma as a trade-off between \textit{unlearning completeness} and \textit{model performance}. 
How to quantitatively calculate the two factors effectively and efficiently without knowing the re-trained model needs to be investigated. 

\subsubsection{Physical Meaning of Power System}

Apart from the completeness and performance trade-off, directly applying the MU algorithms from machine learning community overlooks the physical meaning of power system. In load forecasting, the ultimate goal is to use the forecast load for downstream tasks, such as dispatching the generator. As shown in \cite{vohra2023end, xu2023e2e, donti2017task}, the forecast error mismatches the generator cost deviation such that a highly accurate load forecaster may not result in economic power system operation. Intuitively, we argue that the performance of MU also deviates from the accuracy criterion to the task-aware generator cost. Therefore, the cost of generator needs to be evaluated as the performance criterion when dealing with the completeness-performance trade-off.

\subsection{Contributions}

\begin{figure}
    \centering
    \includegraphics[width=0.75\linewidth]{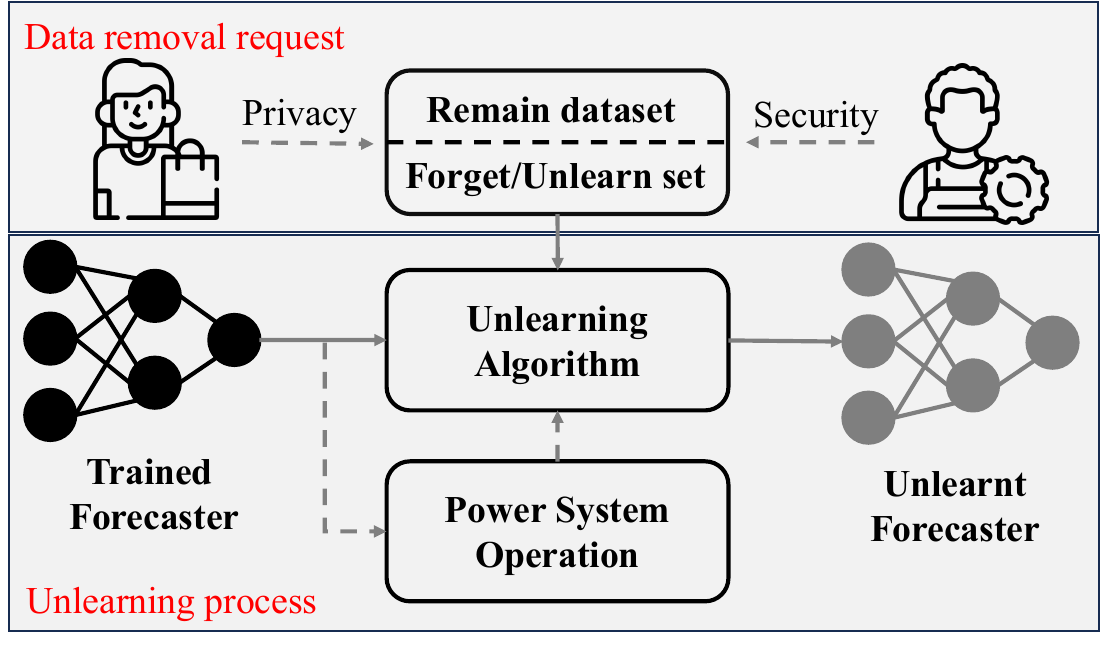}
    \caption{A workflow of machine unlearning. The data removal request can be made by privacy and security concerns. An unlearning algorithm is developed to update the forecaster with the role of power system operation being considered.}
    \label{fig:unlearning_structure}
\end{figure}

The contributions of this paper are summarized as follows:
\begin{itemize}
    \item \textbf{Machine Unlearning}: To our knowledge, this is the first paper applying machine unlearning to power system applications. Specifically, we introduce machine unlearning to the load forecasting model (shown by Fig.\ref{fig:unlearning_structure}). The influence of forget dataset on the trained model is evaluated by influence function-based approach, which is eliminated by Newton's update.
    \item \textbf{Completeness-Performance Trade-off}: We show that complete unlearning can inevitably influence statistical performance of the load forecaster, such as MSE and MAPE. To overcome the dilemma, the influence function is used to quantify the impact on the statistical performance of each sample, which allows reweighting the remaining dataset through optimization and improving performance through performance-aware machine unlearning (PAMU).
    \item \textbf{Task-aware Machine Unlearning}: Finally, we demonstrate that statistical performance cannot reflect the ultimate goal of power system operation, such as minimizing the cost of generator dispatch. Therefore, a task-aware machine unlearning (TAMU) is proposed by formulating the unlearning objective as a trilevel optimization. We theoretically prove the existence of the gradient of such task-aware objective, which is key to sample re-weighting.
\end{itemize}

\section{Machine Unlearning for Load Forecasting}\label{sec:preliminary}

\subsection{Parametric Load Forecasting Model}

In this paper, we consider the load forecasting problem with $n$ loads/participants. Given dataset $\mathcal{D} = \{(\bm{x}^i,\bm{y}^i)\}_{i=1}^N$. Let $\bm{x}^i\in\mathbb{R}^{n\times M}$ be the feature matrix, i.e., each load has feature of length $M$, and $\bm{y}^i\in\mathbb{R}^n$ be the ground truth load. A parametric forecast model $\bm{f}(\cdot;\bm{\theta}):\mathbb{R}^{n\times M}\times\mathbb{R}^{P}\rightarrow\mathbb{R}^{n}$ can be trained as
\begin{equation}\label{eq:ori_obj}
    \bm{\theta}^\star = \arg\min_{\bm{\theta}}\mathcal{L}(\bm{\theta}) = \arg\min_{\bm{\theta}}\frac{1}{N}\sum_{i=1}^N\ell(\bm{f}(\bm{x}^i ; \bm{\theta}),\bm{y}^i)
\end{equation}
where $\mathcal{L}(\bm{\theta})$ is the training loss. For simplicity, denote $\ell(\bm{f}(\bm{x}^i ; \bm{\theta}),\bm{y}^i)$ as $\ell^i(\bm{\theta})$ as the loss on the $i$-th sample. MSE is commonly used as the training loss by assuming that the forecast error follows Gaussian distribution, i.e., $\ell^i(\bm{\theta}) = \|\bm{f}(\bm{x}^i ; \bm{\theta}) - \bm{y}^i\|^2_2$. 

In addition to the training dataset $\mathcal{D}$, there is a test dataset $\mathcal{D}_{\text{test}}$ on which a test criterion can be evaluated on:
\begin{equation}\label{eq:ori_test}
    \mathcal{L}_{\text{test}}(\bm{\theta}^\star) = \frac{1}{N_{\text{test}}}\sum_{i=1}^{N_{\text{test}}}\ell_{\text{test}}^i(\bm{\theta}^\star)
\end{equation}

The test criterion $\ell_{\text{test}}^i(\bm{\theta})$ can be different from the training loss $\ell^i(\bm{\theta})$. For instance, the load forecasting model can be trained with MSE loss but is usually evaluated by MAPE, etc. In this paper, we call the loss/criterion such as MSE and MAPE as statistical(-driven) loss/criterion.

\subsection{Influence Function}

The influence function defines a second-order method to evaluate parameter changes when training samples are up-weighted by a small amount \cite{cook1982residuals}. Define a sub-dataset $\mathcal{D}_{\text{up}} \subseteq \mathcal{D}$. For every sample $j\in\mathcal{D}_{\text{up}}$ up-weighted by $\epsilon^j$, the new objective function can be written as
\begin{equation}\label{eq:mod_obj}
    \begin{aligned}
        \bm{\theta}_{\text{mod}}^\star & = \arg\min_{\bm{\theta}}\mathcal{L}_{\text{mod}}(\bm{\theta}) \\
        & = \arg\min_{\bm{\theta}} \underbrace{\frac{1}{N}\sum_{i\in\mathcal{D}} \ell^i(\bm{\theta})}_{\mathcal{L}(\bm{\theta})} + \underbrace{\frac{1}{N}\sum_{j\in\mathcal{D}_{\text{up}}}\epsilon^j\ell^j(\bm{\theta})}_{\mathcal{L}_{\text{up}}(\bm{\theta})}
    \end{aligned}
\end{equation} 
The first-order optimality condition gives that
\begin{equation}\label{eq:mod_optimality}
    \nabla\mathcal{L}_{\text{mod}}(\bm{\theta}_{\text{mod}}^\star) = \bm{0}
\end{equation}
Apply the first-order Taylor expansion around $\bm{\theta}^\star$ on \eqref{eq:mod_optimality}:
\begin{equation*}
    \nabla \mathcal{L}_{\text{mod}}(\bm{\theta}^\star) + \nabla^2 \mathcal{L}_{\text{mod}}(\bm{\theta}^\star)(\bm{\theta}_{\text{mod}}^\star - \bm{\theta}^\star) \cong \bm{0}
\end{equation*}
Consequently, up-weighting samples in $\mathcal{D}_{\text{up}}$ can approximately result in parameter changes
\begin{equation}\label{eq:mod_parameter_1}
    \bm{\theta}_{\text{mod}}^\star - \bm{\theta}^\star \cong -\left(\nabla^2 \mathcal{L}_{\text{mod}}(\bm{\theta}^\star)\right)^{-1}\nabla \mathcal{L}_{\text{mod}}(\bm{\theta}^\star)
\end{equation}

Furthermore, since $\nabla \mathcal{L}(\bm{\theta}^\star) = \bm{0}$, $\nabla \mathcal{L}_{\text{mod}}(\bm{\theta}^\star) = \nabla \mathcal{L}_{\text{up}}(\bm{\theta}^\star)$. Eq. \eqref{eq:mod_parameter_1} can be rewritten as
\begin{equation}\label{eq:mod_parameter_2}
    \bm{\theta}_{\text{mod}}^\star - \bm{\theta}^\star \cong -\left(\nabla^2 \mathcal{L}_{\text{mod}}(\bm{\theta}^\star)\right)^{-1}\nabla \mathcal{L}_{\text{up}}(\bm{\theta}^\star)
\end{equation}
When $\epsilon_j$ is small and/or $|\mathcal{D}_{\text{up}}| \ll |\mathcal{D}|$, $\nabla^2 \mathcal{L}_{\text{mod}}(\bm{\theta}^\star)\cong \nabla^2 \mathcal{L}(\bm{\theta}^\star)$. Therefore, \eqref{eq:mod_parameter_2} is further approximated as
\begin{equation}\label{eq:mod_parameter_3}
    \bm{\theta}_{\text{mod}}^\star - \bm{\theta}^\star \cong -\left(\nabla^2 \mathcal{L}(\bm{\theta}^\star)\right)^{-1}\nabla \mathcal{L}_{\text{up}}(\bm{\theta}^\star)
\end{equation}
where $\nabla \mathcal{L}_{\text{up}}(\bm{\theta}^\star) = \frac{1}{N}\sum_{j\in\mathcal{D}_{\text{up}}}\epsilon^j\nabla\ell^j(\bm{\theta}^\star)$ and $\nabla^2 \mathcal{L}(\bm{\theta}^\star) = \frac{1}{N}\sum_{j\in\mathcal{D}}\nabla^2\ell^j(\bm{\theta}^\star)$. 

We highlight that \eqref{eq:mod_parameter_1}-\eqref{eq:mod_parameter_3} are Newton's update on the parameter with respect to the new objective \eqref{eq:mod_obj}. Therefore, for the multivariate linear load forecaster with MSE loss, \eqref{eq:mod_parameter_1} and \eqref{eq:mod_parameter_2} are exact updates on the trained model $\bm{\theta}^\star$. 

\subsection{Machine Unlearning Algorithm}


From a data privacy perspective, participants are eligible to ask the SO to remove their data and influence on the trained model $\bm{\theta}^\star$. When a request is made on record $j$, the corresponding datum $(\bm{x}^j,\bm{y}^j)$ needs to be removed from the training dataset. Meanwhile, $\mathcal{D}$ can contain erroneous or malicious data, caused by improper data collection or poisoning attacks, whose influence on the trained forecaster needs to be removed as well. 

Define $\mathcal{D}_{\text{unlearn}}\subset\mathcal{D}$ as the dataset that needs to be removed and $|\mathcal{D}_{\text{unlearn}}| \ll |\mathcal{D}|$. The remaining dataset is denoted as $\mathcal{D}_{\text{remain}}=\mathcal{D}\setminus\mathcal{D}_{\text{unlearn}}$. 
A commonly used MU algorithm can be directly derived from the influence function by setting $\epsilon_j = -1$ in \eqref{eq:mod_obj}. As a result, \eqref{eq:mod_parameter_1} can be modified as
\begin{equation}\label{eq:unlearning}
    \bm{\theta}_{\text{remain}}^\star  
    \cong \bm{\theta}^\star -\left(\sum_{i\in\mathcal{D}_{\text{remain}}}\nabla^2\ell^i(\bm{\theta}^\star)\right)^{-1}\sum_{i\in\mathcal{D}_{\text{remain}}}\nabla\ell^i(\bm{\theta}^\star) \\
\end{equation}
For a linear forecaster, unlearning \eqref{eq:unlearning} is \textbf{complete} as it is guaranteed to converge at $\bm{\theta}_{\text{remain}}^\star$, the model retrained by $\mathcal{D}_{\text{remain}}$. Similar unlearning algorithms can also be derived from \eqref{eq:mod_parameter_2} and \eqref{eq:mod_parameter_3}. 


\section{Performance-aware Machine Unlearning}\label{sec:pumu}

A complete MU algorithm on linear load forecaster such as \eqref{eq:unlearning} can inevitably influence the performance of the test dataset (will be shown in the simulation). Following the previous work in \cite{wu2022puma}, a performance-aware machine unlearning (PAMU) is derived by re-weighting the remaining samples based on their distinct contribution to the statistic criterion \eqref{eq:ori_test}.


To start, the influence function \eqref{eq:mod_parameter_3} can be further extended to assess the performance change of the test set due to the up-weighted objective \eqref{eq:mod_obj} \cite{bae2022if, koh2017understanding}. The performance on the test dataset for model parameterized by $\bm{\theta}^\star_{\text{remain}}$ can be written as
\begin{equation}\label{eq:mod_test}
    \mathcal{L}_{\text{test}}(\bm{\theta}^\star_{\text{remain}}) = \frac{1}{N_{\text{test}}}\sum_{i=1}^{N_{\text{test}}}\ell_{\text{test}}^i(\bm{\theta}_{\text{remain}}^\star)
\end{equation}
Applying first-order Taylor expansion on \eqref{eq:mod_test} gives:
\begin{equation}\label{eq:mod_test_approximation_1}
    \begin{aligned}
        & \mathcal{L}_{\text{test}}(\bm{\theta}^\star_{\text{remain}}) \\
        \cong & \underbrace{\frac{1}{N_{\text{test}}}\sum_{i=1}^{N_{\text{test}}}\ell_{\text{test}}^i(\bm{\theta}^\star)}_{\mathcal{L}_{\text{test}}(\bm{\theta}^\star)} + \frac{1}{N_{\text{test}}}\sum_{i=1}^{N_{\text{test}}}\nabla\ell_{\text{test}}^i(\bm{\theta}^\star)^T(\bm{\theta}_{\text{remain}}^\star - \bm{\theta}^\star) \\
    \end{aligned}
\end{equation}

To eliminate the performance change \eqref{eq:mod_test_approximation_1}, the remaining dataset can be re-weighted. The idea is straightforwardly that, after unlearning, different remaining samples will have different influence on the performance, which needs to be re-weighted as if they are being re-trained.

The new objective function on the re-weighted remaining dataset can be written as
\begin{equation}\label{eq:reweight_obj}
    \bm{\theta}_{\text{remain},\epsilon}^\star = \arg\min_{\bm{\theta}} \frac{1}{N}\sum_{i\in\mathcal{D}_{\text{remain}}} \epsilon^i\ell^i(\bm{\theta})
\end{equation} 
where $\epsilon^i$ is an unknown weight for sample $i$ in the remaining dataset. Referring to \eqref{eq:mod_parameter_1}, the parameter changes can be approximated as
\begin{equation}\label{eq:weight_parameter_1}
    \bm{\theta}_{\text{remain},\epsilon}^\star - \bm{\theta}^\star \cong -\left(\sum_{i\in\mathcal{D}_{\text{remain}}} \epsilon^i\nabla^2\ell^i(\bm{\theta}^\star)\right)^{-1} \sum_{i\in\mathcal{D}_{\text{remain}}} \epsilon^i\nabla\ell^i(\bm{\theta}^\star)
\end{equation}
Plugging \eqref{eq:weight_parameter_1} into \eqref{eq:mod_test_approximation_1}, the performance changes can be written as:
\begin{equation}\label{eq:mod_test_approximation_2}
    \mathcal{L}_{\text{test}}(\bm{\theta}^\star_{\text{remain},\epsilon}) - \mathcal{L}_{\text{test}}(\bm{\theta}^\star) \cong \bm{m}^T \sum_{i\in\mathcal{D}_{\text{remain}}} \epsilon^i\nabla\ell^i(\bm{\theta}^\star)
\end{equation}
where
\begin{equation}\label{eq:mod_matrix}
    \bm{m}^T = -\frac{1}{N_{\text{test}}}\sum_{i \in \mathcal{D}_{\text{test}}}\nabla\ell_{\text{test}}^i(\bm{\theta}^\star) ^T\left(\sum_{i\in\mathcal{D}_{\text{remain}}} \epsilon^i\nabla^2\ell^i(\bm{\theta}^\star)\right)^{-1}
\end{equation}
When $\epsilon^i$ is close to 1, the $\bm{m}$ vector can be approximated as
\begin{equation}\label{eq:mod_matrix_appro}
    \tilde{\bm{m}}^T = -\frac{1}{N_{\text{test}}}\sum_{i\in\mathcal{D}_{\text{test}}}\nabla\ell_{\text{test}}^i(\bm{\theta}^\star) ^T\left(\sum_{i\in\mathcal{D}_{\text{remain}}} \nabla^2\ell^i(\bm{\theta}^\star)\right)^{-1}
\end{equation}

Our goal is to find an optimal weights to improve the test set performance, which can be formulated as a constrained optimization problem:
\begin{equation}\label{eq:alg_unchanged}
    \begin{aligned}
         \bm{\epsilon}^\star = & \arg\min_{\bm{\epsilon}} \tilde{\bm{m}}^T \sum_{i\in\mathcal{D}_{\text{remain}}} \epsilon^i\nabla\ell^i(\bm{\theta}^\star)  \\
        & \text{s.t.}  \quad\quad\quad \frac{1}{N_{\text{remain}}}\|\bm{\epsilon} - \bm{1}\|_1 \leq \lambda_1, \quad \|\bm{\epsilon} - \bm{1}\|_\infty \leq \lambda_\infty
    \end{aligned}
\end{equation}
where $\bm{\epsilon}\in\mathbb{R}^{|\mathcal{D}_{\text{remain}}|}$ and $\bm{1}\in\mathbb{R}^{|\mathcal{D}_{\text{remain}}|}$.

In \eqref{eq:alg_unchanged}, the weights of the remaining samples are optimized so that the influence of forgetting $\mathcal{D}_{\text{unlearn}}$ is reduced.  Since the first-order Taylor expansion \eqref{eq:mod_test_approximation_1} is a local approximation, the 1-norm and inf-norm constraints are added to control aggregated and individual re-weighting. When $\lambda_1\rightarrow0$ or $\lambda_\infty\rightarrow0$, $\bm{\epsilon}\rightarrow1$, representing complete machine unlearning \eqref{eq:unlearning}. When $\lambda_1$ and $\lambda_\infty$ become larger, the performance of the test dataset improves, while the completeness of the unlearning is reduced. Therefore, by controlling $\lambda_1$ and $\lambda_\infty$, the trade-off between MU completeness and performance changes can be balanced. 

Since both $\tilde{\bm{m}}$ and $\nabla\ell^i(\bm{\theta})^\star,i\in\mathcal{D}_{\text{remain}}$ are calculated in advance, \eqref{eq:alg_unchanged} is a convex optimization problem that can be easily solved. Once the optimal weights $\bm{\epsilon}^\star$ are optimized, we can unlearn $\mathcal{D}_{\text{unlearn}}$ through \eqref{eq:weight_parameter_1}. Regarding different choices of $\ell_{\text{test}}$, e.g. MSE and MAPE, the remaining dataset can be reweighted in distinct manners. In addition, it is also possible to integrate different criteria.

Furthermore, compared to \cite{wu2022puma}, the objective of \eqref{eq:alg_unchanged} does not take the absolute value. Therefore, the objective of \eqref{eq:alg_unchanged} can be negative and it is allowed to improve performance beyond the originally trained model. Any uncovered biased data in $\mathcal{D}_{\text{remain}}$ will be assigned a smaller weight, and unlearning becomes a one-step continual learning on the re-weighted samples. Re-weighting the samples to improve the model performance has been used in other load forecasting algorithm \cite{wang2023improving}. However, we directly find the suitable weights on the trained parameter $\bm{\theta}^\star$ through an optimization problem \eqref{eq:alg_unchanged} and update the model using the second-order approach \eqref{eq:weight_parameter_1}.

\section{Task-aware Machine Unlearning}\label{sec:tapumu}

\subsection{Formulation and Algorithm}

In power systems, the forecast load is further used to schedule generators, and the statistic accuracy of the forecast load is eventually converted into the deviation of the generator cost, which is strongly linked to the value of each sample as well as the profit of the SO and participants. As a result, PAMU guided by the statistic-driven criterion may not reflect on the ultimate goal of power system operation, and a further step on PAMU is needed to balance the generator cost, which can be done by taking the generator cost as the new test criterion.

To measure the impact of model parameter $\bm{\theta}$ on the operation cost, the following task-aware criterion $\mathcal{L}_{\text{gen}}(\bm{\theta})$ can be formulated:
\begin{equation}\label{eq:obj_spo}
    \begin{aligned}
        & \min_{\bm{\theta}} \frac{1}{N_{\text{test}}} \sum_{i=1}^{N_{\text{test}}} \ell_{\text{gen}}^i(\bm{\theta}) \\
        \text{s.t.} & \left\{
                \begin{array}{l}
                        \text{(Re-dispatch):} \\
                        (\bm{P}_{ls}^{i\star}, \bm{P}_{gs}^{i\star}) \in \arg\min \{  c_{ls2}\|\bm{P}_{ls}^i\|_2^2 + c_{gs2}\|\bm{P}_{gs}^i\|_2^2 \\
                        \qquad \qquad \qquad \qquad + c_{ls}\|\bm{P}_{ls}^i\|_1 +  c_{gs}\|\bm{P}_{gs}^i\|_1: \\
                        \qquad \qquad \qquad \qquad (\bm{P}_{ls}^i, \bm{P}_{gs}^i) \in \mathcal{C}_{\text{redispatch}}(\bm{P}_g^{i\star},\bm{y}^i) \} \\
                        \text{(Dispatch):} \\
                        \bm{P}_g^{i\star} \in \arg\min \{\bm{P}_g^{iT}\bm{Q}_g\bm{P}_g^i + \bm{c}_g^T\bm{P}_g^{i} + c_{ls}\|\bm{s}^i\|_1: \\
                        \qquad \qquad \qquad \qquad (\bm{P}_g^i, \bm{s}^i) \in \mathcal{C}_{\text{dispatch}}(\hat{\bm{y}}^i) \} \\
                        \text{(Forecast):} \\
            \hat{\bm{y}}^i = \bm{f}(\bm{x}^i;\bm{\theta})
                \end{array}\right. \\
                & \qquad \text{for all } i = 1,\cdots,|\mathcal{D}_{\text{test}}|
    \end{aligned}
\end{equation}

Detailed formulations can be found in Appendix \ref{app:operation}.
The task-aware criterion \eqref{eq:obj_spo} can be viewed as a trilevel optimization problem with two lower levels, taking the expectation over the test dataset. For each sample, the lower level one is a dispatch problem that minimizes the generator cost subject to the system operation constraint $\mathcal{C}_{\text{dispatch}}$. $\bm{P}_g$ is the generator dispatch. Lower level two is a re-dispatch problem which aims to balance any under- or over-generation due to inaccurate forecast through load shedding $\bm{P}_{ls}$ and generation storage $\bm{P}_{gs}$, under the constraint set $\mathcal{C}_{\text{redispatch}}$. The upper level, which represents the expected operation cost, can be determined as the integration of the two stages:
\begin{equation}\label{eq:gen_cost}
    \begin{array}{rl}
        &\ell_{\text{gen}}(\bm{P}_g, \bm{P}_{ls}, \bm{P}_{gs};\bm{\theta}) \\
    = &\bm{P}_g^T\bm{Q}_g\bm{P}_g + c_{ls2}\|\bm{P}_{ls}\|_2^2 + c_{gs2}\|\bm{P}_{gs}\|_2^2 \\ 
      &  + \bm{c}_g^T\bm{P}_g +  c_{ls1}\|\bm{P}_{ls}\|_1 + c_{gs1}\|\bm{P}_{gs}\|_1
    \end{array}
\end{equation}
 When $\bm{\theta}$ is fixed and if each lower-level problem has a unique optimum, \eqref{eq:obj_spo} is the expected real-time power system operation cost on the test dataset. 

Referring to \eqref{eq:mod_test_approximation_1}, to evaluate the influence on the generator cost, the gradient $\nabla \ell^i_{\text{gen}} (\bm{\theta}^\star)$ needs to be calculated, which seems to be a problem due to the nested structure and constraints in \eqref{eq:obj_spo}. To solve the problem, firstly, for each sample $i$, it can be observed that the lower level problems are sequentially connected, that is, the input to stage one problem is the forecast load while the input to stage two problem is the generator dispatch status from stage one. Second, the lower-level problems are also independent among samples and the constraints. Therefore, the lower-level optimizations can be viewed as composite function for each sample. Let $\bm{P}_{g}^i = \bm{p}^\star_1(\hat{\bm{y}}^i)$ and $\bm{P}_{ls,gs}^i = \bm{p}^\star_2(\bm{P}_{g}^i,\bm{y}^i)$ be the optimal solution map for dispatch and re-dispatch, the individual generator cost \eqref{eq:gen_cost} can be written as a composite function:
\begin{equation}\label{eq:composite_spo}
    \ell_{\text{gen}}^i(\bm{\theta}^\star) = \ell_{\text{gen}}^i(\bm{p}^\star_1(\hat{\bm{y}}^i), \bm{p}^\star_2(\bm{p}^\star_1(\hat{\bm{y}}^i), \bm{y}^i))
\end{equation}
Alternatively, we can view the lower-level optimizations as sequential layers upon the parametric forecasting model. The layer, which represents a constrained optimization problem, is named as differentiable convex layer \cite{agrawal2019differentiable}. 

Consequently, TAMU can be achieved by replacing the statistic metric $\ell_{\text{test}}^i(\bm{\theta}^\star)$ by $\ell_{\text{gen}}^i(\bm{\theta}^\star)$, followed by finding the weights of the remaining dataset \eqref{eq:alg_unchanged} and updating the parameters by \eqref{eq:weight_parameter_1}. 

\begin{figure}
    \centering
    \includegraphics[width=0.9\linewidth]{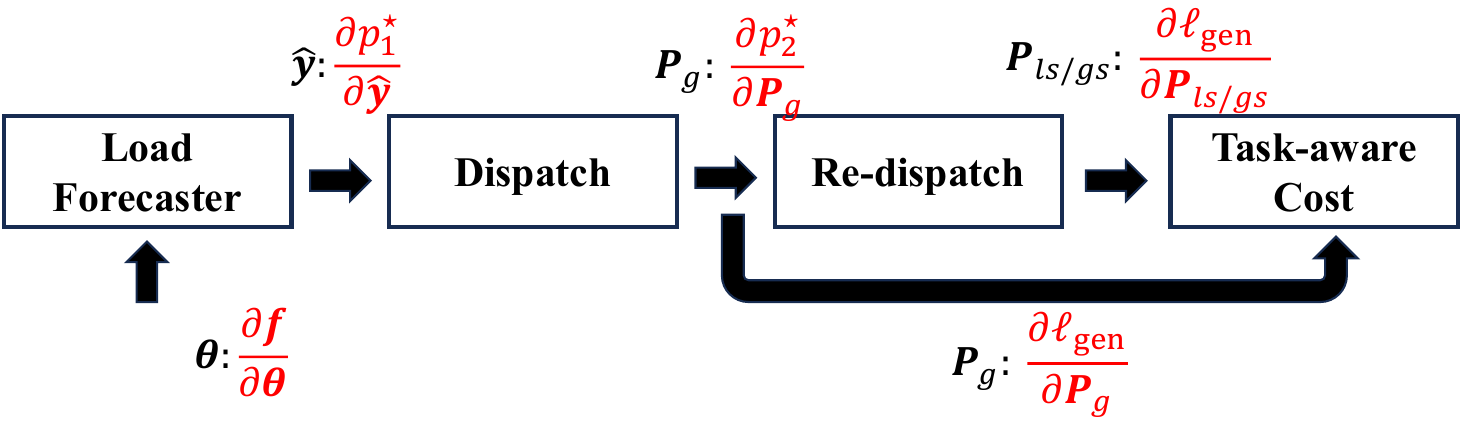}
    \caption{The structure of tri-level optimization \eqref{eq:obj_spo} viewed as layers in the forward pass. The gradients used in \eqref{eq:cost_chain_rule} are highlighted in red.}
    \label{fig:gradient_flow}
\end{figure}

The last issue that needs to be resolved is to calculate the gradient of \eqref{eq:composite_spo} as required by \eqref{eq:mod_test_approximation_1}. From the chain rule, the gradient of \eqref{eq:composite_spo} can be written as:
\begin{equation}\label{eq:cost_chain_rule}
    \begin{aligned}
        & \quad\frac{\partial \ell^i_{\text{gen}}(\bm{\theta}^\star)}{\partial \bm{\theta}} \\
        = & \left(\frac{\partial \ell^i_{\text{gen}}(\bm{P}_g^i, \bm{P}_{ls,gs}^i)}{\partial \bm{P}_g} + \frac{\partial \ell^i_{\text{gen}}(\bm{P}_g^i, \bm{P}_{ls,gs}^i)}{\partial \bm{P}_{ls,gs}} \frac{\partial \bm{p}_2^\star(\bm{P_g}^i)}{\partial \bm{P}_g} \right) \\
        & \quad \times \frac{\partial \bm{p}_1^\star(\hat{\bm{y}}^i)}{\partial \hat{\bm{y}}}\frac{\partial \bm{f}(\bm{\theta}^\star)}{\partial \bm{\theta}}
    \end{aligned}
\end{equation}
with the gradient flow highlighted in Fig.\ref{fig:gradient_flow}. In \eqref{eq:cost_chain_rule}, the gradient $\partial \ell^i_{\text{gen}}(\bm{\theta}^\star) \slash \partial \bm{\theta} $ exists if the gradients through the differentiable convex layers, namely $\partial \bm{p}_1^\star(\hat{\bm{y}}^i)\slash \partial \hat{\bm{y}}$ and $\partial \bm{p}_2^\star(\bm{P_g}^i) \slash \partial \bm{P}_g$, exist, which is fulfilled under some assumptions in the following proposition. 

\begin{proposition}\label{theorem:gradient}
    The gradients $\partial \bm{p}_1^\star(\hat{\bm{y}}^i)\slash \partial \hat{\bm{y}}$ and $\partial \bm{p}_2^\star(\bm{P_g}^i) \slash \partial \bm{P}_g$ exist, which do not depend on $\hat{\bm{y}}^i$ and $\bm{P_g}^i$, respectively, if 1). $\bm{Q}$ is positive definite, $\bm{c}_{ls2}$ and $\bm{c}_{gs2}$ are positive; and 2). The linear independent constraint qualification (LICQ) is satisfied at the optimum of each of the lower-level problems.
\end{proposition}

The proof can be found in Appendix \ref{app:qp_gradient}. 

\subsection{Extension to Neural Network based Load Forecaster}

The unlearning algorithm \eqref{eq:unlearning} is complete on the linear load forecaster as the training objective is quadratic. However, this condition is usually not satisfied for neural networks. In the meantime, its Hessian can be singular due to early stop of training. This makes the influence function approximate poorly to the parameter and performance changes \cite{bae2022if}, and it becomes harder to evaluate the trade-off between unlearning completeness and model performance in PAMU and TAMU.

To address this problem, we assume that there exists a load forecasting model which is not trained by the consumers' data in the service provided by SO. The model can be a pre-trained model which is publicly available or can be trained on historic non-sensitive data by the SO. Using the idea of transfer learning \cite{zhuang2020comprehensive}, the SO can then use the pre-trained load forecaster as a deep feature extractor and use the consumers' data to fine tune the last layer. Therefore, only the last layer needs to be unlearnt.

\begin{figure}
    \centering
    \includegraphics[width=0.95\linewidth]{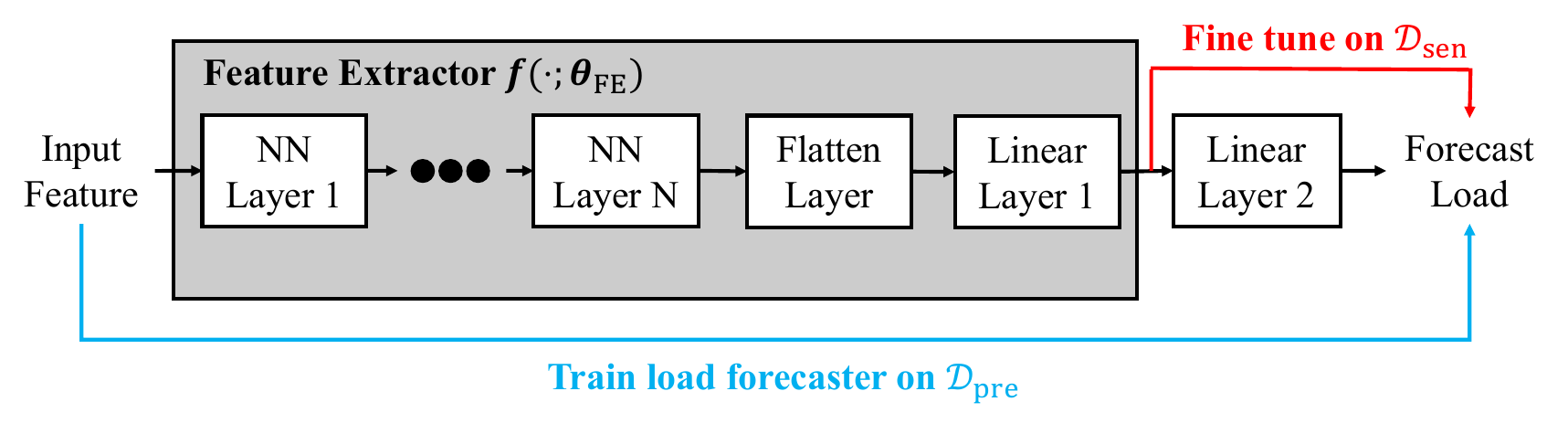}
    \caption{Structure of NN based load forecaster and feature extractor. All the layers except for the Flatten Layer and the Linear Layer 2 contain activations.}
    \label{fig:feature_extrator}
\end{figure}

Practically, we first divide the training dataset into pre-trained and user-sensitive data as $\mathcal{D}_{\text{pre}}$ and $\mathcal{D}_{\text{sen}}$, respectively, with $\mathcal{D}_{\text{pre}} \cap \mathcal{D}_{\text{sen}} = \emptyset$. The pre-trained data is assumed to be collected neutrally, which does not violate any participant's privacy and is error-free, while the user-sensitive data may not be. We then pre-train a load forecaster on the $\mathcal{D}_{\text{pre}}$ using regular stochastic gradient descent (SGD), and the trained model (except for the last layer) can be used as a deep feature extractor $\bm{f}(\cdot;\bm{\theta}^\star_{\text{FE}})$. As illustrated in Fig.\ref{fig:feature_extrator}, $\mathcal{D}_{\text{sen}}$ is further used to fine-tune Linear Layer 2 by the MSE loss. Since using a stochastic gradient method can introduce uncertainties, we propose to fine-tune the last layer analytically on $\mathcal{D}_{\text{sen}}$ according to the following proposition.

\begin{proposition}\label{theorem:dense}
    The optimization problem of minimizing the MSE loss on a linear layer without activations is quadratic and has unique minimizer if the extracted features from $\bm{f}(\cdot;\bm{\theta}^\star_{\text{FE}})$ are linearly independent.
\end{proposition}

The proof can be found in Appendix \ref{app:proof_fine_tune}. 

According to Proposition \ref{theorem:dense}, \texttt{ReLU} activation cannot be used in Linear Layer 1 as it can result in trivial output when the extracted features are negative for some of the samples in the remaining dataset. When an unlearning is requested, the same unlearning algorithms developed previously can be applied on Linear Layer 2 alone and MU \eqref{eq:unlearning} is complete. Since the feature extractor trains only on the pre-train data, it does not contain sensitive information that needs to be unlearnt.

\subsection{Computations}

In this section, we discuss some computational issues and some useful open-source packages for the developed unlearning algorithms.

\subsubsection{Inversion of Hessian}

Machine unlearning \eqref{eq:unlearning} and calculation of vector $\tilde{\bm{m}}$ in PAMU and TAMU require matrix inversion of the Hessian matrix. In general, second-order differentiation on training loss is time consuming, as storing and inverting the Hessian matrix requires $\mathcal{O}(d^3)$ operations, where $d$ represents the number of parameters in the load forecast model. 

Using $\tilde{\bm{m}}$ \eqref{eq:mod_matrix_appro} as a example:
\begin{equation}
    \tilde{\bm{m}}^T = \underbrace{-\frac{1}{N_{\text{test}}}\sum_{i=1}^{N_{\text{test}}}\nabla\ell_{\text{test}}^i(\bm{\theta}^\star) ^T}_{\bm{v}^T\in\mathbb{R}^{1\times d}}
    \underbrace{\left(\sum_{i\in\mathcal{D}_{\text{remain}}} \nabla^2\ell^i(\bm{\theta}^\star)\right)^{-1}}_{\bm{H}^{-1}\in\mathbb{R}^{d\times d}}
\end{equation}

Calculating $\tilde{\bm{m}}$ can be reduced to solve a linear system:
\begin{equation}\label{eq:linear_system}
    \bm{H} \cdot \tilde{\bm{m}} = \bm{v}
\end{equation}

The conjugate gradient (CG) descent algorithm can be applied to solve \eqref{eq:linear_system} up to $d$ iterations. We also apply the Hessian vector product (HVP) \cite{pearlmutter1994fast} to directly calculate $\bm{H} \tilde{\bm{m}}^k$ for the $k$-th iteration in CG so that the Hessian matrix will never be explicitly calculated and stored. HVP is computationally efficient as it only requires one modified forward and backward pass. Similarly, to implement PAMU or TAMU, we can modify the objective directly into the sum of training loss weighted by $\bm{\epsilon}^\star$ from \eqref{eq:alg_unchanged} and implement the same CG and HVP procedure. We implement these functionalities using a modified version of \texttt{Torch-Influence} package \cite{bae2022if}. 

\subsubsection{Differentiable Convex Layer}

In TAMU, the gradient of generator cost \eqref{eq:cost_chain_rule} can be analytically written according to Proposition \ref{theorem:app_qp_affine} in Appendix \ref{app:qp_gradient}. It also requires the forward pass to solve the dispatch and re-dispatch problems. In the simulation, we model the operation problems and \eqref{eq:alg_unchanged} by \texttt{Cvxpy} \cite{diamond2016cvxpy}. When calculating the gradient, we use \texttt{PyTorch} automatic differentiation package and \texttt{CvxpyLayers} \cite{agrawal2019differentiable} to implement fast batched forward and backward passes. 

\section{Experiments and Results}\label{sec:simulation}

\subsection{Simulation Settings}

We use an open-source dataset from the Texas Backbone Power System \cite{lu2023synthetic} which includes meteorological and calendar features and loads in 2019 with a resolution of one hour. The dispatch and re-dispatch problems are solved on a modified IEEE bus-14 system to demonstrate the proposed algorithms. Three parametric load forecasting models, namely multivariate linear regression, convolutional neural network (CNN), and MLP-Mixer \cite{tolstikhin2021mlp}, are trained by MSE loss. Detailed experimental settings can be found in Appendix \ref{app:exp_setting}.

\subsection{Unlearning Performance on the Linear Model}

\subsubsection{Unlearning Performance}

\begin{figure*}[h]
     \centering
     \begin{subfigure}[b]{0.3\linewidth}
         \centering
         \includegraphics[width=\textwidth]{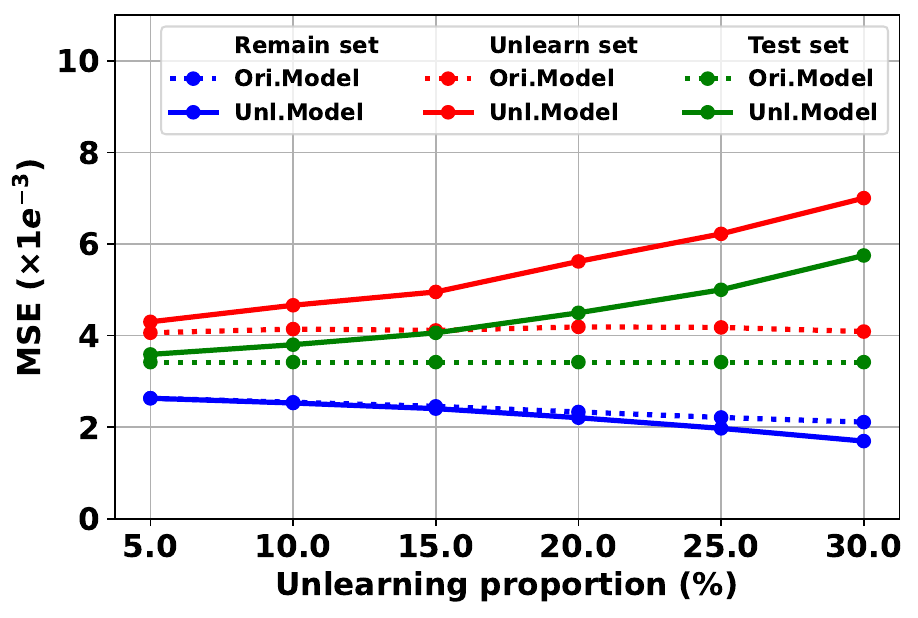}
         \caption{MSE}
         \label{fig:unlearn_performance_mse}
     \end{subfigure}
     \hfill
     \begin{subfigure}[b]{0.3\linewidth}
         \centering
         \includegraphics[width=\textwidth]{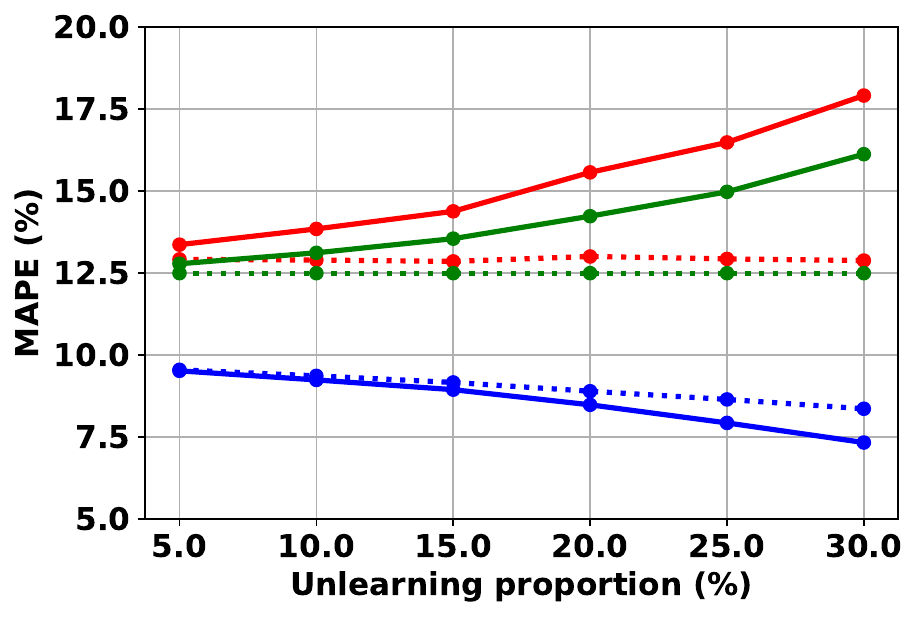}
         \caption{MAPE}
         \label{fig:unlearn_performance_mape}
     \end{subfigure}
     \hfill
     \begin{subfigure}[b]{0.3\linewidth}
         \centering
         \includegraphics[width=\textwidth]{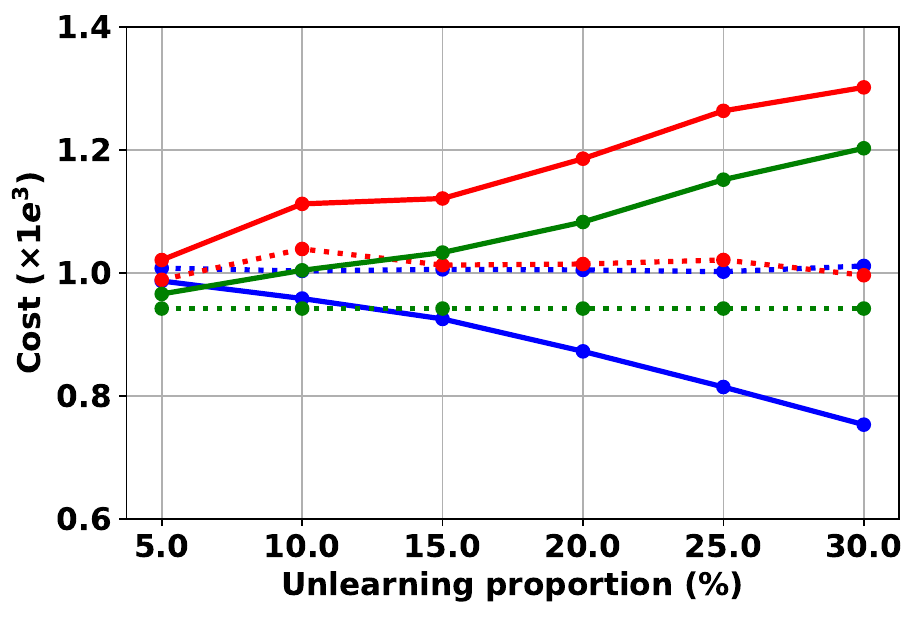}
         \caption{Cost}
         \label{fig:unlearn_performance_cost}
     \end{subfigure}
        \caption{Performance of complete machine unlearning algorithm \eqref{eq:unlearning} on remain (blue), unlearn (red) and test dataset (blue) of the \textbf{linear} load forecaster. The dotted curves report the performance of the original model and the solid curves are the performance of the unlearnt model.}
        \label{fig:unlearn_performance}
\end{figure*}

Unlearning performances on the linear load forecasting model under various unlearning criteria are summarized in Fig.\ref{fig:unlearn_performance}. We have verified that the unlearning algorithm \eqref{eq:unlearning} results in the same updated parameter as the one re-trained on the remaining dataset under all unlearning rates.

Note that the dotted curves, which represent the performance of the original model, only slightly change over the various unlearning ratios. Broadly speaking, the performance gaps between the unlearnt and original models becomes larger as the unlearning ratio increases. Especially, all the performance criteria on the test dataset become worse when the unlearning proportion increases, which verifies the statement that unlearning can inevitably degrade the generalization ability of the trained model. For instance, the generator cost can increase by 20\% when 20\% of the training data are unlearnt. In contrast, the performance of the remain dataset improves as the unlearning ratio increases. This is because when the original model is unlearnt, the model parameters are updated and fitted more on the remaining dataset. Moreover, it can be observed that the trends of performance changes of the unlearnt model are distinct for different criterion. In detail, the generator cost (Fig.\ref{fig:unlearn_performance_cost}) diverges more significantly from the original model, compared to MSE and MAPE. 

\subsubsection{Performance Sensitivity Analysis}

\begin{figure*}
     \centering
     \begin{subfigure}[b]{0.3\linewidth}
         \centering
         \includegraphics[width=\textwidth]{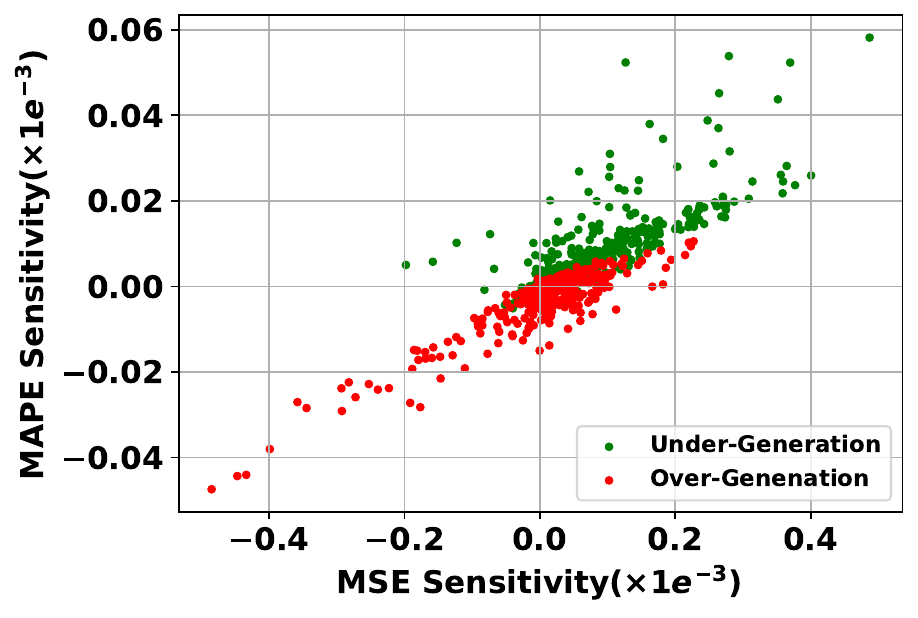}
         \caption{MSE and MAPE ($\bm{r=0.829}$)}
         \label{fig:sensitivity_mse-mape}
     \end{subfigure}
     \hfill
     \begin{subfigure}[b]{0.3\linewidth}
         \centering
         \includegraphics[width=\textwidth]{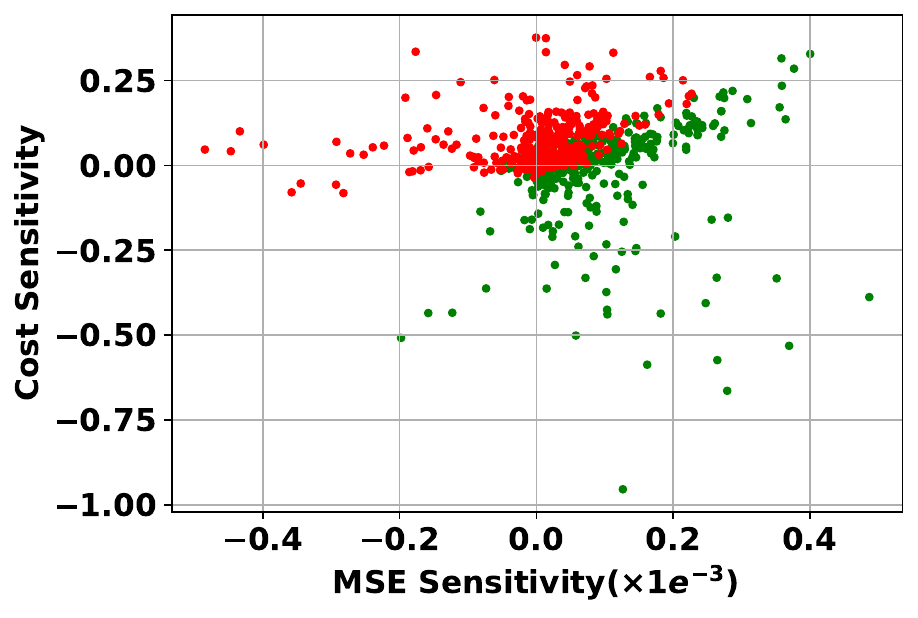}
         \caption{MSE and Cost ($\bm{r=0.073}$)}
         \label{fig:sensitivity_mse-cost}
     \end{subfigure}
     \hfill
     \begin{subfigure}[b]{0.3\linewidth}
         \centering
         \includegraphics[width=\textwidth]{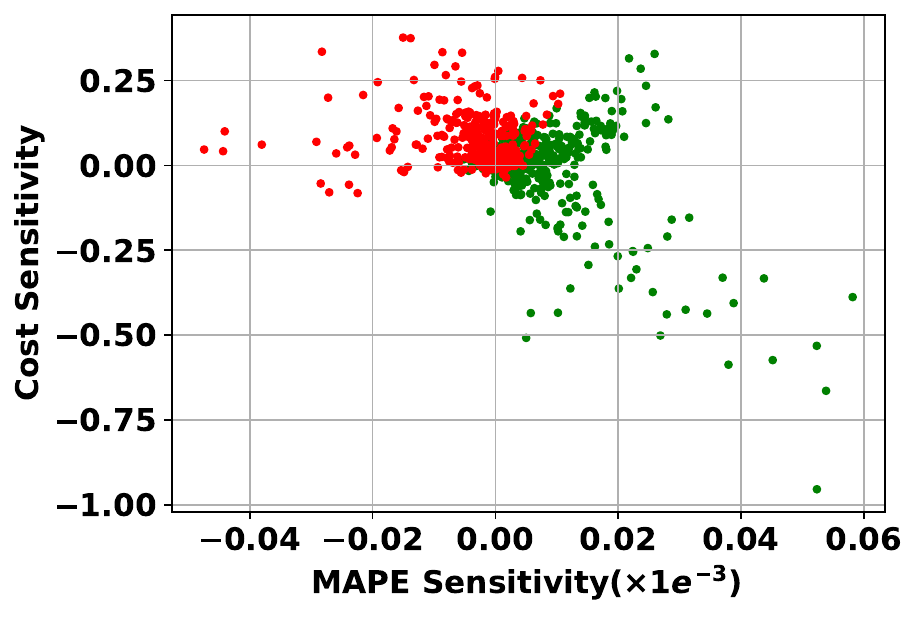}
         \caption{MAPE and Cost ($\bm{r=-0.480}$)}
         \label{fig:sensitivity_mape-cost}
     \end{subfigure}
        \caption{Relationship on the influences of MSE, MAPE, and Cost criteria on the test dataset from the samples in remain dataset. The $r$ values are Pearson correlation coefficients.}
        \label{fig:sensitivity}
\end{figure*}

For each sample in the remaining dataset, we can calculate its influence on the expected performance of the test dataset. The remaining dataset is chosen as it is re-weighted by PAMU and TAMU. For $i\in\mathcal{D}_{\text{remain}}$, the influence can be found by \eqref{eq:mod_test_approximation_2} and \eqref{eq:mod_matrix_appro} with $\epsilon_i=1$, i.e.,
\begin{equation}
    \begin{aligned}
        & \mathcal{I}_{\text{test}}^i \\
        = & -\frac{1}{N_{\text{test}}}\sum_{j\in\mathcal{D}_{\text{test}}}\nabla\ell_{\text{test}}^j(\bm{\theta}^\star) ^T\left(\sum_{j\in\mathcal{D}_{\text{remain}}} \nabla^2\ell^j(\bm{\theta}^\star)\right)^{-1} \nabla\ell^i(\bm{\theta}^\star)
    \end{aligned}
\end{equation}
where the test loss $\ell_{\text{test}}(\cdot)$ can be MSE, MAPE or Cost \eqref{eq:composite_spo}. To visualize the relationship among these criteria, we randomly draw 1k samples with equal size of under- and over-generation cases from the remaining dataset. For each sample, the under-generation means that the sum of the forecast loads is lower than the sum of the ground-truth load, and the over-generation is opposite. The relationships of any two of the criteria are illustrated in Fig.\ref{fig:sensitivity} with associated Pearson correlation coefficients (the $r$ value) calculated. Since the performance changes are modeled linearly by first-order Taylor expansion \eqref{eq:mod_test_approximation_1} and the objective of re-weighting optimization is also linear \eqref{eq:alg_unchanged}, Pearson correlation coefficient is a suitable indicator of the linear relationship. Using MSE and MAPE as an example, the Pearson correlation coefficient is defined as
\begin{equation}
    r_{\text{MSE}, \text{MAPE}} = \frac{\sum_{i} (\mathcal{I}_{\text{MSE}}^i - \bar{\mathcal{I}}_{\text{MSE}}) (\mathcal{I}_{\text{MAPE}}^i - \bar{\mathcal{I}}_{\text{MAPE}})}{\sqrt{\sum_{i} (\mathcal{I}_{\text{MSE}}^i - \bar{\mathcal{I}}_{\text{MSE}})^2} \sqrt{\sum_{i} (\mathcal{I}_{\text{MAPE}}^i - \bar{\mathcal{I}}_{\text{MAPE}})^2} }
\end{equation}
where $\bar{\mathcal{I}}_{\text{MSE}}$ and $\bar{\mathcal{I}}_{\text{MAPE}}$ are the average of MSE and MAPE influence, respectively.

In Fig.\ref{fig:sensitivity}, positive sensitivity represents the degradation of performance after unlearning such sample. That is, after this sample is unlearnt, the MSE, MAPE, or average generator cost on test dataset increases. First, the Pearson correlation coefficients have clearly demonstrated that there exists a strong positive linear relationship between the two statistic criteria (0.829), while this relationship is insignificant between the statistic and task-aware criteria (0.073 between MSE and Cost and -0.480 between MAPE and Cost). These distinct relationships imply that balancing performance by one statistical criterion is likely effective on the other. In contrast, balancing the performance by statistical criteria can unlikely be effective on the generator cost and vice versa. Secondly, as the under-generation is more costly than the over-generation, unlearning an under-generation sample tends to reduce the overall generator cost with negative sensitivities. As shown by Fig.\ref{fig:sensitivity_mse-cost} and Fig.\ref{fig:sensitivity_mape-cost}, if the sensitivities are projected to the y-axis, most of the negative sensitivities are contributed by the under-generation samples, which verifies our intuition. However, it does not occur in MSE and MAPE as they are almost centrally symmetric around the origin in Fig.\ref{fig:sensitivity_mse-mape}.

The above discussions can verify the intuition that the statistic performance cannot reflect and may even conflict with the task-aware operation cost.

\subsubsection{Performances of PAMU and TAMU}

\begin{figure*}
     \centering
     \begin{subfigure}[b]{0.3\linewidth}
         \centering
         \includegraphics[width=\textwidth]{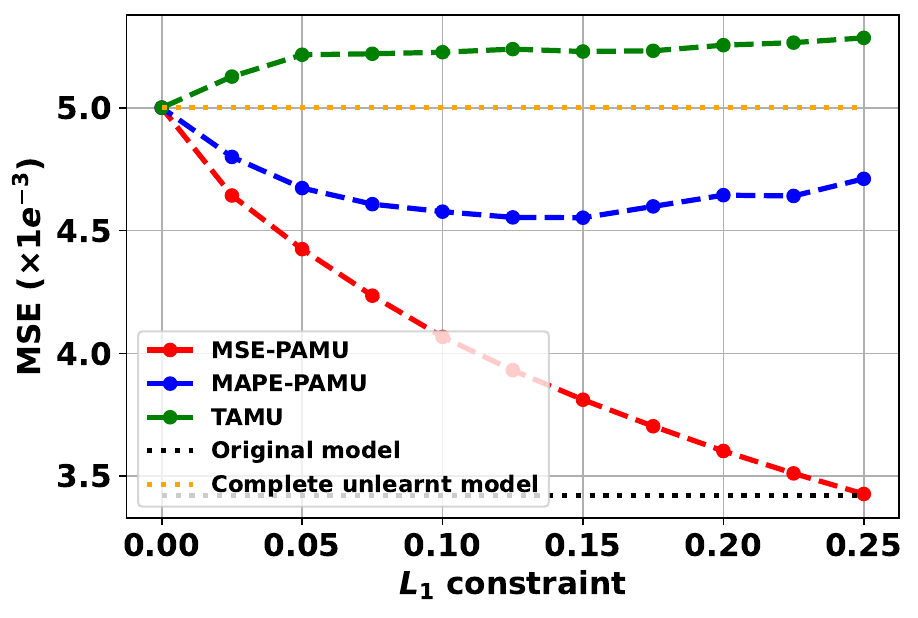}
         \caption{MSE}
         \label{fig:unchange_mse}
     \end{subfigure}
     \hfill
     \begin{subfigure}[b]{0.3\linewidth}
         \centering
         \includegraphics[width=\textwidth]{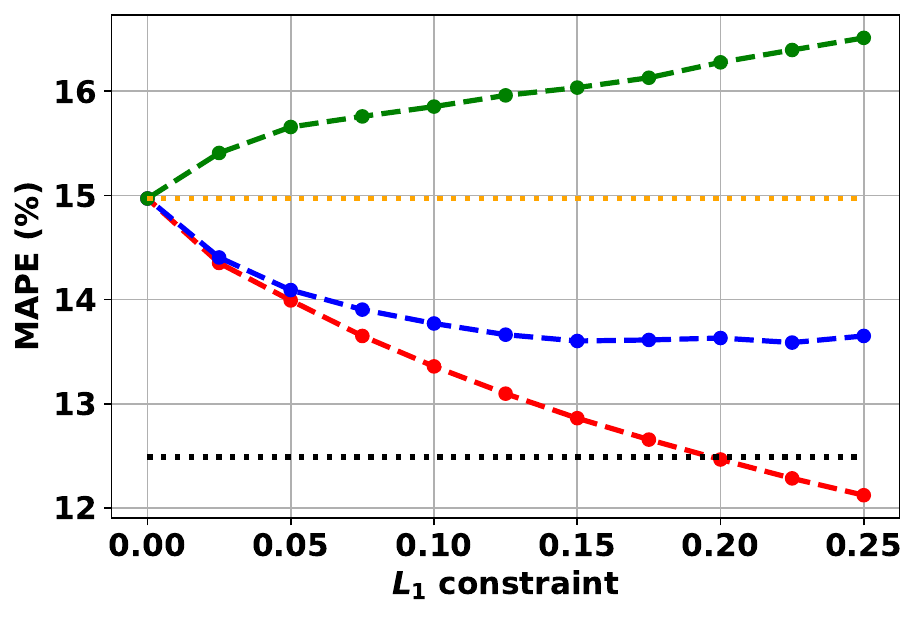}
         \caption{MAPE}
         \label{fig:unchang_mape}
     \end{subfigure}
     \hfill
     \begin{subfigure}[b]{0.3\linewidth}
         \centering
         \includegraphics[width=\textwidth]{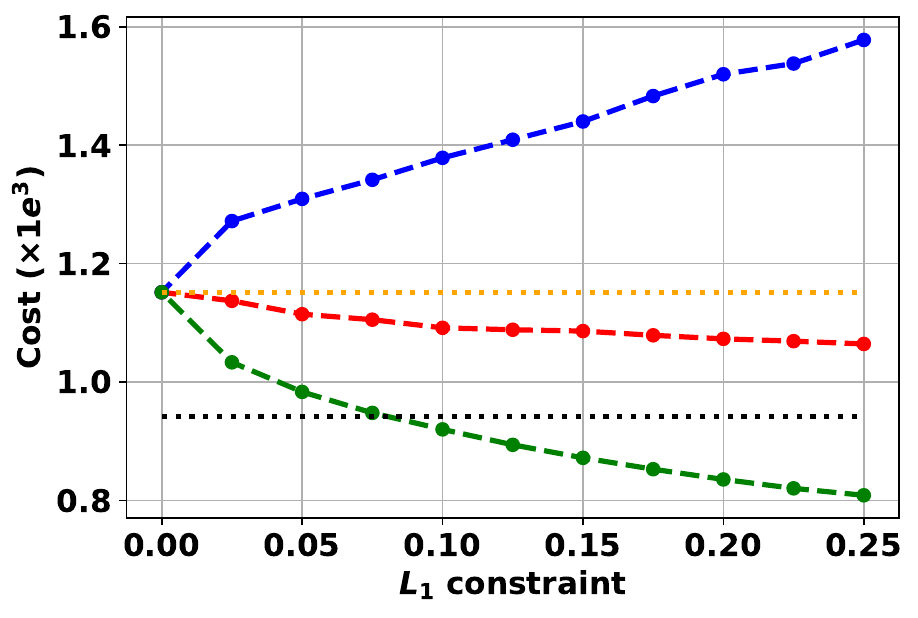}
         \caption{Cost}
         \label{fig:unchange_cost}
     \end{subfigure}
        \caption{Performance of PAMU and TAMU with different test criteria. 
        a), b), and c) are performances on the test dataset evaluated by MSE, MAPE, and average generator cost, respectively. The performance of the original model and the model unlearnt by complete unlearning \eqref{eq:unlearning} are represented by the black and orange lines, respectively.}
        \label{fig:unchange}
\end{figure*}

\begin{figure}
    \centering
    \includegraphics[width=0.6\linewidth]{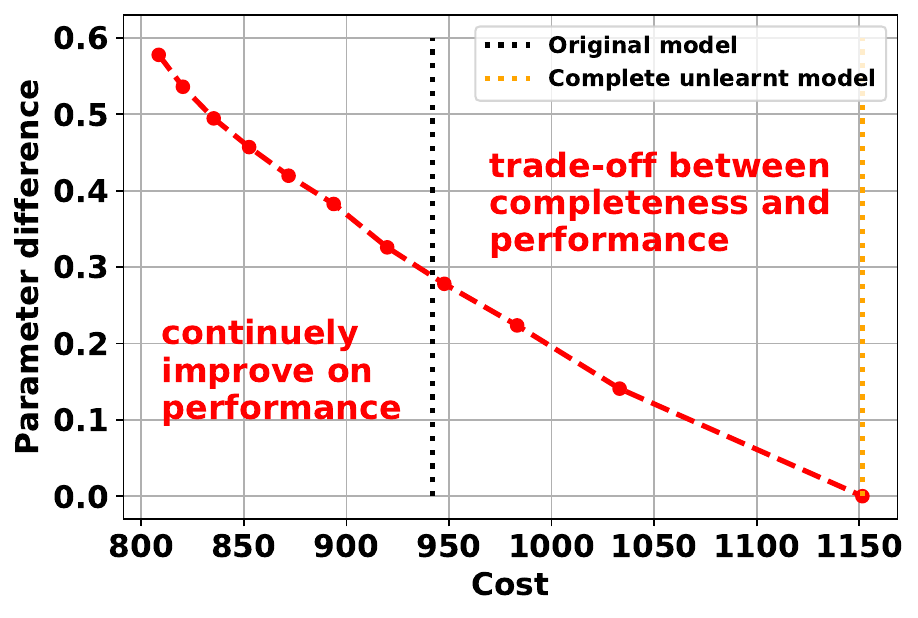}
    \caption{Trade-off between MU completeness and the operation cost.}
    \label{fig:trade_off}
\end{figure}

The performance of PAMU and TAMU on the test dataset is reported in Fig.\ref{fig:unchange} in which 25\% training data is removed. To balance the trade-off, $\lambda_1$ is varied and the inf-norm constraint $\lambda_\infty$ in \eqref{eq:alg_unchanged} is set as 1. That is, the weight of a remaining sample can very from 0 to 2. First, unlearning by balancing one of the criteria can effectively maintain the performance of the same criterion (e.g., red curve in Fig.\ref{fig:unchange_mse}, blue curve in Fig.\ref{fig:unchang_mape}, and green curve in Fig.\ref{fig:unchange_cost}). When $\lambda_1$ approaches 0, the PAMU and TAMU become complete with the same performance as the retrained model in all criteria, as no samples can be re-weighted. When $\lambda_1$ increases, the performance of the original model is recovered and the divergence to the retrained model increases. After $\lambda_1$ is further increased, better performance is achieved, resulting in a new type of continual learning through sample re-weighting. As a result, the proposed PAMU and TAMU can effectively balance the completeness and performance trade-off in MU by changing $\lambda_1$. In addition, Fig.\ref{fig:trade_off} illustrates the parameter difference to the retrained model (evaluated by 2-norm) vs the generator cost, which clearly demonstrates the trade-off as well.

Meanwhile, it is observed that the cost curves perform differently compared to the MSE and MAPE curves. When balancing the cost, both MSE and MAPE get worse. In contrast, balancing the MSE can also keep/improve the MAPE performance to some extent, and vice versa. This observation is in line with the analysis on the Pearson correlation coefficient in the previous section. 


\subsection{Unlearning Performance on the NN Forecaster}

Since the fine-tuning objective on $\mathcal{D}_{\text{sen}}$ is quadratic (by Proposition \ref{theorem:dense}), the direct unlearning is also complete for the NN load forecaster. We can expect that the unlearning behaviors are similar to the linear counterpart. Therefore, we only highlight some of the simulation results and leave details in Appendix \ref{app:experiment}. 
Similarly to the linear counterpart, unlearning part of the training dataset can deteriorate the performance of the test set. The performance-aware unlearning algorithm can effectively balance the unlearning completeness and model performance, which is more effective on the criterion it is evaluated on. However, instead of having opposite statistical and cost trends in Fig.\ref{fig:unchange}, all criteria are improved with decay speed. This is because the NN-based load forecaster is more accurate than the linear counterpart, resulting in a less significant misalignment between the load forecast accuracy and the generator cost.

\section{Conclusion}\label{sec:conclusion}

This paper introduces machine unlearning algorithm for load forecasting model to eliminate the influence of data that is adversarial or contains sensitive information of individuals. The influence function provides a theoretical foundation that is further used to evaluate the impact of unlearning on the performance of the test dataset. A performance aware machine unlearning is proposed by re-weighting the remaining dataset. To handle the divergence between statistical and task-aware criteria, we propose task-aware machine unlearning. The simulation results verify that the proposed task-aware algorithm can significantly reduce the generator cost on the test dataset by compensating for the unlearning completeness.

\appendix 
\numberwithin{equation}{section}
\setcounter{equation}{0}

\subsection{Power System Operation Models}\label{app:operation}

In the US, the following network constrained economic dispatch (NCED) problem is widely adopted \cite{conejo2018power}. Given the forecast load $\hat{\bm{y}}$ on each bus,
\begin{equation*}
    \begin{aligned}
        (\bm{P}_g^\star, \bm{\vartheta}^\star, \bm{s}^\star) = & \arg\min_{\bm{P}_g, \bm{\vartheta}, \bm{s}} \bm{P}_g^T\bm{Q}_g\bm{P}_g + \bm{c}_g^T\bm{P}_g + c_{ls}\|\bm{s}\|_1 \\
        \text{s.t. } & \underline{\bm{P}}_{g} \leq \bm{P}_g \leq \bar{\bm{P}}_g \\
                    & \bm{B}_{\text{bus}}\bm{\vartheta} = \bm{C}_g\bm{P}_g - \bm{C}_l(\hat{\bm{y}} - \bm{s}) \\
                    & \underline{\bm{P}}_f \leq \bm{B}_f\bm{\vartheta} \leq \bar{\bm{P}}_f \\
                    & \bm{s} \geq 0, \quad \bm{\vartheta}_{\text{ref}} = 0
    \end{aligned}
\end{equation*}
In the dispatch problem, a quadratic generator cost is adopted. In addition, $\bm{B}_{\text{bus}}$ and $\bm{B}_f$ are the bus susceptance and branch succeptance matrices, respectively. $\bm{C}_g$ and $\bm{C}_l$ are the generator and load incidence matrices, respectively. A slack variable $\bm{s}\geq 0$ with large cost $c_{ls}$ is introduced to ensure feasibility.

After the generators are scheduled, any over- and/or under- generations are penalized when the actual load $\bm{y}$ is realized in real time. In detail, given $\bm{P}_g^\star$, we consider the following optimization problem modified from \cite{zhang2022cost}:
\begin{equation*}
    \begin{aligned}
        (\bm{P}_{ls}^\star, \bm{P}_{gs}^\star, \bm{\vartheta}^\star) = & \arg\min_{\bm{P}_{ls}, \bm{P}_{gs}, \bm{\vartheta}}  c_{ls2}\|\bm{P}_{ls}\|_2^2 + c_{gs2}\|\bm{P}_{gs}\|_2^2 \\ &  \quad \quad + c_{ls1}\|\bm{P}_{ls}\|_1 + c_{gs1}\|\bm{P}_{gs}\|_1 \\
        \text{s.t. }
                    & \bm{B}_{\text{bus}}\bm{\vartheta} = \bm{C}_g(\bm{P}_g^\star - \bm{P}_{gs})- \bm{C}_l({\bm{y}} - \bm{P}_{ls}) \\
                    & \underline{\bm{P}}_f \leq \bm{B}_f\bm{\vartheta} \leq \bar{\bm{P}}_f \\
                    & \bm{P}_{ls} \geq 0, \quad \bm{P}_{gs} \geq 0, \quad \bm{\vartheta}_{\text{ref}} = 0
    \end{aligned}
\end{equation*}
where $\bm{P}_{ls}$ and $\bm{P}_{gs}$ are the load shedding and generation storage. The second-order cost $c_{gs2} < c_{ls2}$ and linear cost $c_{gs1} < c_{ls1}$ are set to penalize more on the load shedding.

\subsection{Proof to Proposition \ref{theorem:gradient}}\label{app:qp_gradient}

We prove Proposition \ref{theorem:gradient} by proving a more general Proposition \ref{theorem:app_qp_affine}. To start, consider the following QP:
\begin{equation}\label{eq:app_qp}
    \begin{aligned}
        \bm{x}^\star  = & \arg\min_{\bm{x}} \frac{1}{2}\bm{x}^T\bm{Q}\bm{x}+\bm{q}^T\bm{x} \\
        \text{s.t. } & \bm{A}\bm{x} + \bm{b} + \bm{g}(\bm{z}) \leq \bm{0} \\
        & \bm{C}\bm{x} + \bm{d} + \bm{h}(\bm{z}) = \bm{0}
    \end{aligned}
\end{equation}
where $\bm{x}\in\mathbb{R}^n$, $\bm{Q}\in\mathbb{R}^{n\times n}$, $\bm{q}\in\mathbb{R}^{n}$, $\bm{A}\in\mathbb{R}^{m\times n}$, $\bm{b}\in\mathbb{R}^{m}$, $\bm{C}\in\mathbb{R}^{p\times n}$, $\bm{d}\in\mathbb{R}^{p}$, and $\bm{z}\in\mathbb{R}^{q}$. $\bm{g}:\mathbb{R}^q\rightarrow\mathbb{R}^m$ and $\bm{h}:\mathbb{R}^q\rightarrow\mathbb{R}^p$ are functions on $\bm{z}$, representing the perturbation parameters. Apart from the linear parametric inequality constraints in \eqref{eq:obj_spo}, we also include the linear parametric term $\bm{g}(\bm{z})$ in the inequality constraint for generalization purposes (and it also gives the same conclusion to Proposition \ref{theorem:gradient}). We call \eqref{eq:app_qp} \textit{affine-parametric}, since the parametric terms $\bm{g}(\bm{z})$ and $\bm{h}(\bm{z})$ are affine in the inequality and equality constraints.


\begin{proposition}\label{theorem:app_qp_affine}
    Given an affine parametric QP \eqref{eq:app_qp}, the optimal primal and dual pair $(\bm{x}^\star, \bm{\lambda}^\star, \bm{\nu}^\star)$ is an affine function of the parameter $(\bm{g}(\bm{z}), \bm{h}(\bm{z}))$ if 1). $\bm{Q}$ is positive definite; and 2). the linear independent constraint qualification (LICQ) is satisfied at $(\bm{x}^\star, \bm{\lambda}^\star, \bm{\nu}^\star)$.
\end{proposition}

\begin{proof}

First, the LICQ states that the gradient of the active constraints (including all equality constraints and active inequality constraints) are linearly independent \cite{jorge2006numerical}. Therefore, $\bm{C}$ is full row rank. Second, the equality Karush–Kuhn–Tucker (KKT) conditions \cite{jorge2006numerical} can be denoted as:
\begin{equation*}
    \bm{G}(\bm{x}^\star,\bm{\lambda}^\star,\bm{\nu}^\star, \bm{z}) = \left(\begin{array}{c}
         \bm{Q}\bm{x}^\star + \bm{q} + \bm{A}^T\bm{\lambda}^\star +  \bm{C}^T\bm{\nu}^\star  \\
          \operatorname{diag}(\bm{\lambda}^\star)(\bm{A}\bm{x}^\star + \bm{b} + \bm{g}(\bm{z})) \\
          \bm{C}\bm{x}^\star + \bm{d} + \bm{h}(\bm{z}) 
    \end{array} \right) = \bm{0}
\end{equation*} 

We divide the proofs by the existence of active constraints.

When there are no active inequality constraints, $\bm{\lambda}^\star = \bm{0}$ due to complementary slackness. Since $\bm{Q}$ is positive definite, the stationary condition gives $\bm{x}^\star = -\bm{Q}^{-1}(\bm{q}+\bm{C}^T\bm{\nu}^\star)$. From the equality constraint, it can be derived that $\bm{\nu}^\star = (\bm{C}\bm{Q}^{-1}\bm{C}^T)^{-1}(-\bm{C}\bm{Q}^{-1}\bm{q} + \bm{d} + \bm{h}(\bm{z}))$. Note that $\bm{C}\bm{Q}^{-1}\bm{C}^T$ is positive definite (thus invertible). Let $\hat{\bm{C}} = \bm{Q}^{-1}\bm{C}^T(\bm{C}\bm{Q}^{-1}\bm{C}^T)^{-1}$, the analytical form for $\bm{x}^\star$ can be written as
\begin{equation}\label{eq:affine_inactive}
    \bm{x}^\star = (-\bm{Q}^{-1}+\hat{\bm{C}}\bm{C}\bm{Q}^{-1})\bm{q}- \hat{\bm{C}}(\bm{d} + \bm{h}(\bm{z}))
\end{equation}
which is affine in $\bm{h}(\bm{z})$.

When there exist some active inequality constraints, let $\tilde{{\bm{\lambda}}}$, $\tilde{\bm{A}}$, $\tilde{\bm{b}}$, and $\tilde{\bm{g}}(\bm{z})$ be the sub-matrices whose rows are indexed by the active constraints. Therefore, $\bm{A}^T\bm{\lambda}^\star = \tilde{\bm{A}}^T\tilde{\bm{\lambda}}^\star$ and the active inequality constraint becomes:
\begin{equation}\label{eq:app_proof_1}
    \tilde{\bm{A}}\bm{x}^\star + \tilde{\bm{b}} + \tilde{\bm{g}}(\bm{z}) = 0
\end{equation}
Since $\bm{Q}$ is positive definite, the stationary condition gives that
\begin{equation}\label{eq:app_proof_2}
    \bm{x}^\star = -\bm{Q}^{-1}(\bm{q} + \tilde{\bm{A}}^T\tilde{\bm{\lambda}}^\star + \bm{C}^T\bm{\nu}^\star)
\end{equation}
Plugging \eqref{eq:app_proof_2} into \eqref{eq:app_proof_1} and the equality condition gives the following matrix form:
\begin{equation}\label{eq:app_proof_3}
    \bm{\mathcal{Q}}
    \left(\begin{array}{c}
        \tilde{\bm{\lambda}}^\star \\
        \bm{\nu}^\star
    \end{array}\right) = \underbrace{\left(\begin{array}{c}
         - \tilde{\bm{A}}\bm{Q}^{-1}\bm{q} + \tilde{\bm{b}} + \tilde{\bm{g}}(\bm{z}) \\
         -\bm{C}\bm{Q}^{-1}\bm{q} + \bm{d} + \bm{h}(\bm{z})
    \end{array}\right)}_{\bm{r}(\bm{z})}
\end{equation}
where
\begin{equation*}
    \begin{aligned}
        \bm{\mathcal{Q}} & 
        = \left(\begin{array}{cc}
        \tilde{\bm{A}}\bm{Q}^{-1}\tilde{\bm{A}}^T & \tilde{\bm{A}}\bm{Q}^{-1}\bm{C}^T \\
        \bm{C}\bm{Q}^{-1}\tilde{\bm{A}}^T & \bm{C}\bm{Q}^{-1}\tilde{\bm{C}}^T
    \end{array}\right) \\ & 
    = \left(\begin{array}{c}
        \tilde{\bm{A}} \\
        \bm{C}
    \end{array}\right)\bm{Q}^{-1}
    \left(
    \begin{array}{cc}
        \tilde{\bm{A}}^T & \bm{C}^T
    \end{array}
    \right)
    \end{aligned}
\end{equation*}
Due to LICQ, $(\tilde{\bm{A}}^T,\bm{C}^T)$ is full column rank. Therefore, $\bm{\mathcal{Q}}$ is positive definite and from \eqref{eq:app_proof_3}
\begin{equation}\label{eq:app_proof_4}
    \left(\begin{array}{c}
        \tilde{\bm{\lambda}}^\star \\
        \bm{\nu}^\star
    \end{array}\right) = \bm{\mathcal{Q}}^{-1}\bm{r}(\bm{z})
\end{equation}
which is affine in $(\tilde{\bm{g}}(\bm{z})^T,\bm{h}(\bm{z})^T)^T$.
Consequently, plugging \eqref{eq:app_proof_4} into \eqref{eq:app_proof_2} gives
\begin{equation}\label{eq:affine_active}
    \bm{x}^\star = -\bm{Q}^{-1}\left(\bm{q} + (\tilde{\bm{A}}^T,\bm{C}^T)\bm{\mathcal{Q}}^{-1}\bm{r}(\bm{z})\right)
\end{equation}
which is affine in $(\tilde{\bm{g}}(\bm{z})^T,\bm{h}(\bm{z})^T)^T$.

\end{proof}

Since the optimal solution of every QP satisfying Proposition \ref{theorem:app_qp_affine} is an affine function of the parameter (\eqref{eq:affine_inactive} and \eqref{eq:affine_active}), the gradients of the convex layers in the dispatch and re-dispatch problems exist and can be analytically written regardless of the perturbed parameter.

Note that with the final representation, the optimal solution $\bm{x}^\star$ still needs to be computed in the forward pass as $\tilde{\bm{A}}$ can only be determined when $\bm{x}^\star$ is known.

\subsection{Proof to Proposition \ref{theorem:dense}}\label{app:proof_fine_tune}

Let $\bm{f}(\cdot;\bm{\theta}_\text{FE}^\star)$ be the trained feature extractor on the pre-train dataset. Let $\bm{X}_{\text{sen}}\in\mathbb{R}^{N_{\text{sen}}\times d}$ be the extracted feature of $\mathcal{D}_{\text{sen}}$ as input to the Linear Layer 2. $N_{\text{sen}}$ is the number of user sensitive data and $d$ is the output size of feature extractor. Note that $d\ll N_{\text{sen}}$ and $\bm{X}_{\text{sen}}$ is full column rank by the condition. Meanwhile, let $\bm{Y}_{\text{sen}}\in\mathbb{R}^{N_{\text{sen}} \times n}$ be the ground truth load over $n$ participants. The parameter of Linear Layer 2 is denoted as $\bm{\Theta}\in\mathbb{R}^{d\times n}$.

Let $\bm{y}_{\cdot, i}\in\mathbb{R}^{N_{\text{sen}}}$ and $\bm{\theta}_{\cdot,i}\in\mathbb{R}^{d}$ be the $i$-th column of $\bm{Y}_{\text{sen}}$ and $\bm{\Theta}$, respectively. The fine-tuning objective can be written as
\begin{equation}\label{eq:fine_tune_obj}
    \mathcal{L}(\bm{\theta}) = \frac{1}{N_{\text{sen}}\cdot n} \sum_{i=1}^{n}\|\bm{y}_{\cdot, i} - \bm{X}_{\text{sen}}\bm{\theta}_i\|_2^2
\end{equation}
Now define $\hat{\bm{X}}_{\text{sen}} = \text{diag}([\underbrace{\bm{X}_{\text{sen}},\cdots,\bm{X}_{\text{sen}}}_{n}]) \in \mathbb{R}^{N_{\text{sen}}n \times dn} $ as a block diagonal matrix packed by $n$ $\bm{X}_{\text{sen}}$s. $\hat{\bm{Y}}_{\text{sen}} = [\bm{y}_{\cdot, 1}^T,\cdots,\bm{y}_{\cdot, n}^T]^T\in\mathbb{R}^{N_{\text{sen}}n}$ and $\hat{\bm{\Theta}} = [\bm{\theta}_{\cdot, 1}^T,\cdots,\bm{\theta}_{\cdot, n}^T]^T\in\mathbb{R}^{dn}$ be the flattened version of $\bm{Y}_{\text{sen}}$ and $\bm{\Theta}$, respectively. It can be verified that \eqref{eq:fine_tune_obj} is equivalent to
\begin{equation}\label{eq:fine_tune_obj_1}
    \mathcal{L}(\bm{\theta}) = \frac{1}{N_{\text{sen}}\cdot n} \left( \hat{\bm{\Theta}}^T \hat{\bm{X}}_{\text{sen}}^T\hat{\bm{X}}_{\text{sen}} \hat{\bm{\Theta}} - 2 \bm{Y}_{\text{sen}}^T \hat{\bm{X}}_{\text{sen}} \hat{\bm{\Theta}} + \hat{\bm{Y}}_{\text{sen}}^T\hat{\bm{Y}}_{\text{sen}} \right)
\end{equation}

Since $\bm{X}_{\text{sen}}$ is full column rank, $\hat{\bm{X}}_{\text{sen}}^T\hat{\bm{X}}_{\text{sen}}$ is positive definite. Therefore, \eqref{eq:fine_tune_obj_1} and \eqref{eq:fine_tune_obj} are quadratic with unique global minimizer.

\subsection{Detailed Experiment Settings}\label{app:exp_setting}

\subsubsection{Data Description}

The meteorological features in the Texas Backbone Power System \cite{lu2023synthetic} include temperature (k), long-wave radiation (w / m2), short-wave radiation (w / m2), zonal wind speed (m / s), meridional wind speed (m / s) and wind speed (m / s), which are normalized according to their individual mean and standard deviation. The calendar feature includes the cosine and sin of the weekday in a week and the hour in a day according to their individual period. Therefore, a single datum is $(\bm{x}^i, \bm{y}^i)\in\mathbb{R}^{14\times 10}\times \mathbb{R}^{14}$. 
We also normalize the target load by its mean and std. Meanwhile, we use the first 80\% data as training dataset and the remaining as test dataset. Finally, the IEEE bus-14 system is modified from \texttt{PyPower}.

\subsubsection{Linear Load Forecaster}

The linear load forecaster can be found by
\begin{equation*}
    \min_{\bm{\theta}}\frac{1}{N\cdot 14}\sum_{i=1}^N\|\bm{x}^i\bm{\theta} - \bm{y}^i\|_2^2
\end{equation*}
where $\bm{\theta}\in\mathbb{R}^{10}$. The quadratic objective can be solved analytically or by using conjugate gradient descent.

\subsubsection{Convolutional NN Load Forecaster}

A CNN is used as feature extractor, which is summarized in Table.\ref{tab:forecast_cnn}.

\begin{table}[h]
    \centering
    \footnotesize
    \caption{Structure of the CNN load forecasting model. For the convolutional layer, $(k: w\times h + s + p)$ represents the $k$ number of filters, kernel size $w\times h$ with $s$ stride and $p$ padding in both sides. For the linear layer, the number indicates the output size. The activation function is written in bracket. }
    \begin{tabular}{c|c}\hline
       Conv Layer 1 & 8: 3 $\times$ 3 + 1 + 1 (ReLU)  \\\hline
       Conv Layer 2 & 8: 4 $\times$ 4 + 2 + 1 (ReLU) \\\hline
       Linear Layer 1  & 64 (tanh) \\\hline
       Linear Layer 2 & 14 (No activation)  \\\hline
    \end{tabular}
    
    \label{tab:forecast_cnn}
\end{table}

\subsubsection{MLP-Mixer Load Forecaster}

MLP-Mixer only contains two types of multi-layer perceptrons (MLPs), which iteratively capture the information on the feature patches and across the feature patches. In our load forecast setting, it iteratively captures the features within each load and across each load. Regardless of its simple structure, it has been reported that MLP-Mixer can have a performance comparable to CNN or attention-based networks, e.g., transformers \cite{tolstikhin2021mlp}. The structure of MLP-Mixer is summarized in Table \ref{tab:forecast_mlpmixer}.

\begin{table}[h]
    \centering
    \footnotesize
    \caption{Structure of the MLP-Mixer load forecasting model with exact settings in \cite{tolstikhin2021mlp}. One basic Mixer block contains two MLP blocks. Each MLP block contains two linear layers and one activation function between them. We also apply layer norms before each MLP block and pooling layers wherever necessary.}
    \begin{tabular}{c|c}\hline
       No. of Patches & 2 \\\hline
       No. of Mixer Blocks & 2 \\\hline
       MLP & 64 (GeLU) \\\hline
       Linear Layer 1  & 64 (tanh) \\\hline
       Linear Layer 2 & 14 (no activation)  \\\hline
    \end{tabular}
    \label{tab:forecast_mlpmixer}
\end{table}

\subsubsection{Traing Configuration}

We use the same training specifications for CNN and MLP-Mixer. We select the first 30\% in the training dataset as the pre-train dataset and the remaining as the user-sensitive dataset. The NN forecaster is trained with 100 epochs, batch size of 16, Adam optimizer with learning rate of $10^{-4}$ and cosine annealing. We also use early stop and record the model with the best performance.

\subsection{Extra Experiment Results}\label{app:experiment}

\begin{figure*}
     \centering
     \begin{subfigure}[b]{0.3\linewidth}
         \centering
         \includegraphics[width=\textwidth]{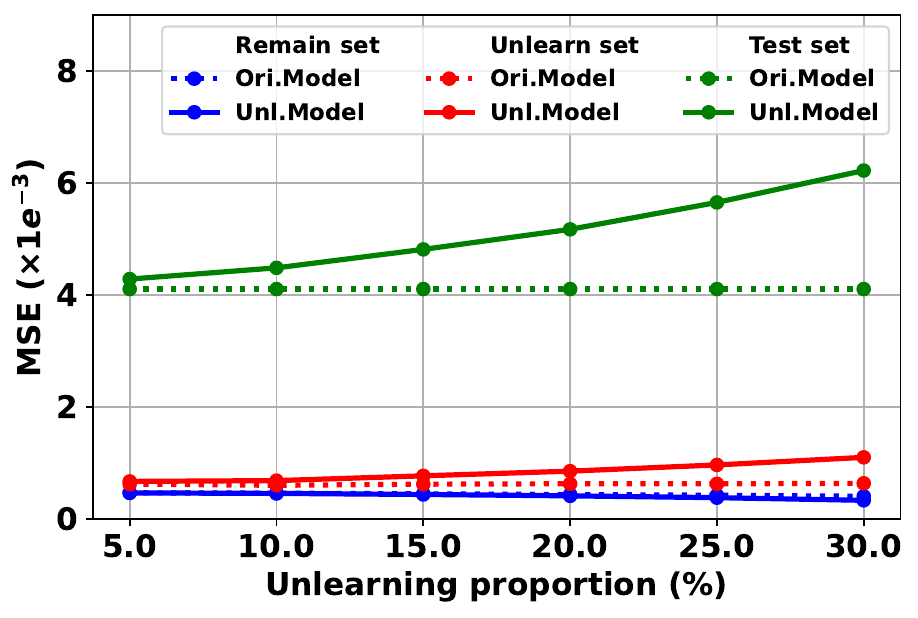}
         \label{fig:nn_conv_unlearn_performance_mse}
     \end{subfigure}
     \hfill
     \begin{subfigure}[b]{0.3\linewidth}
         \centering
         \includegraphics[width=\textwidth]{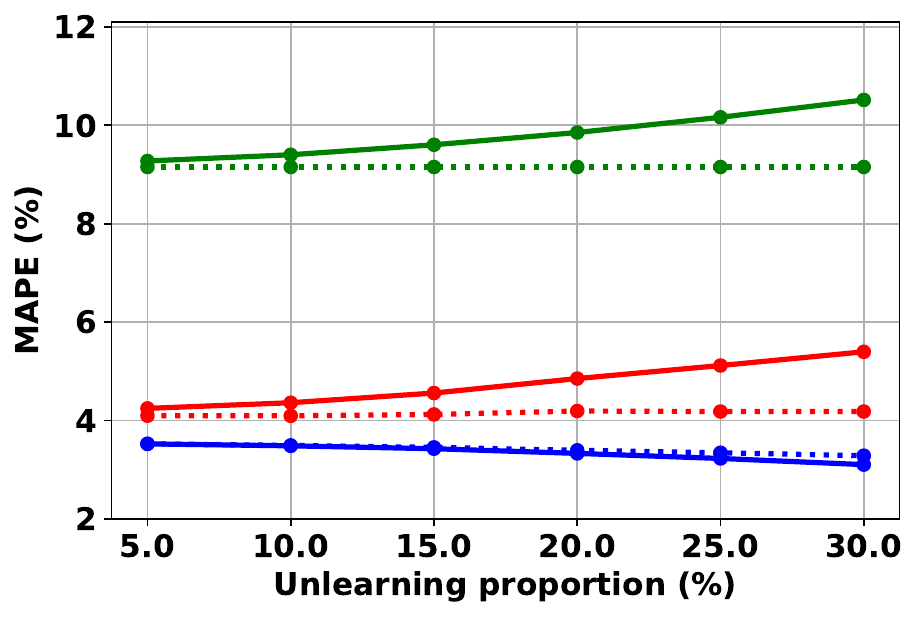}
         \label{fig:nn_conv_unlearn_performance_mape}
     \end{subfigure}
     \hfill
     \begin{subfigure}[b]{0.3\linewidth}
         \centering
         \includegraphics[width=\textwidth]{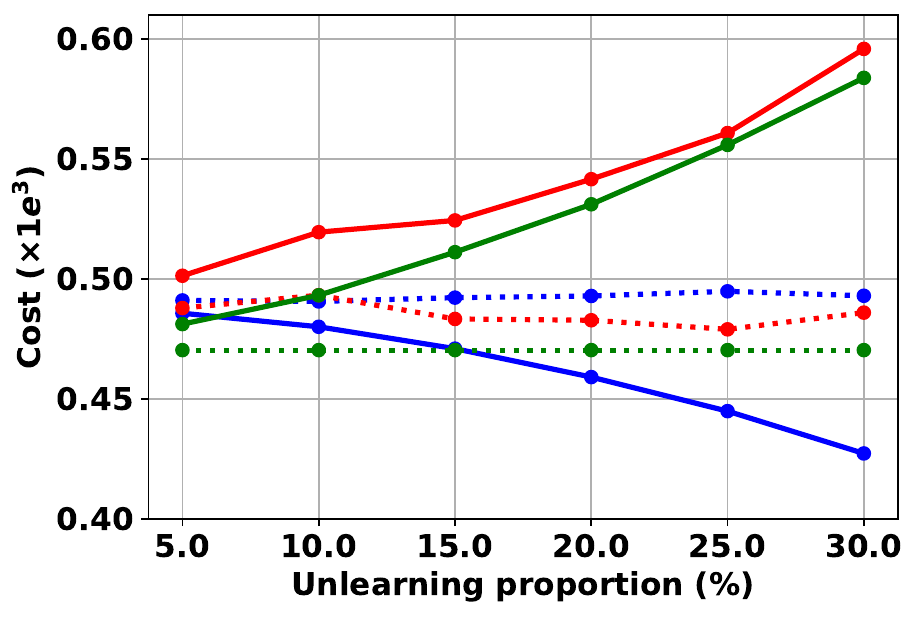}
         \label{fig:nn_conv_unlearn_performance_cost}
     \end{subfigure}
     \hfill
     \begin{subfigure}[b]{0.3\linewidth}
         \centering
         \includegraphics[width=\textwidth]{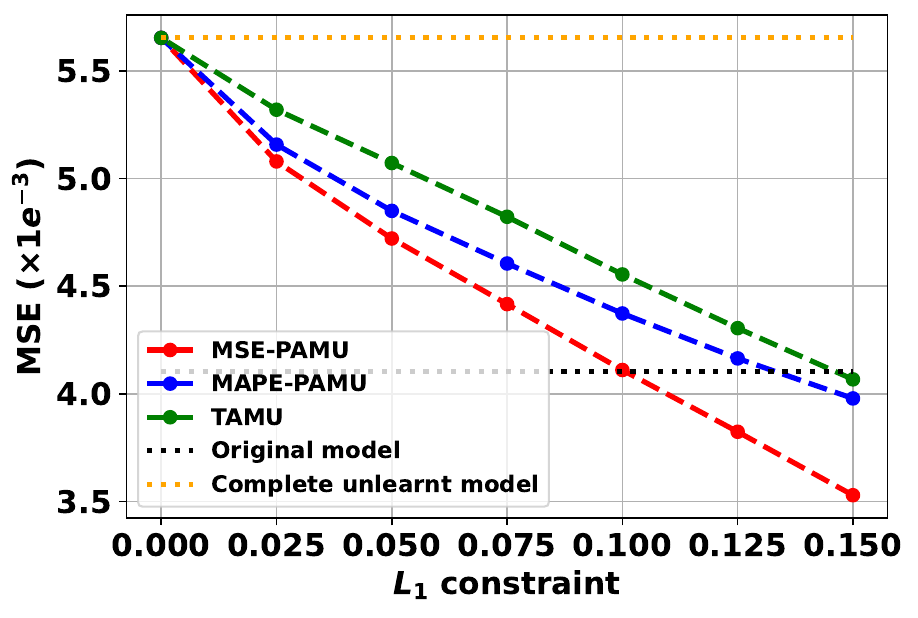}
         \label{fig:nn_conv_unchange_mse}
     \end{subfigure}
     \hfill
     \begin{subfigure}[b]{0.3\linewidth}
         \centering
         \includegraphics[width=\textwidth]{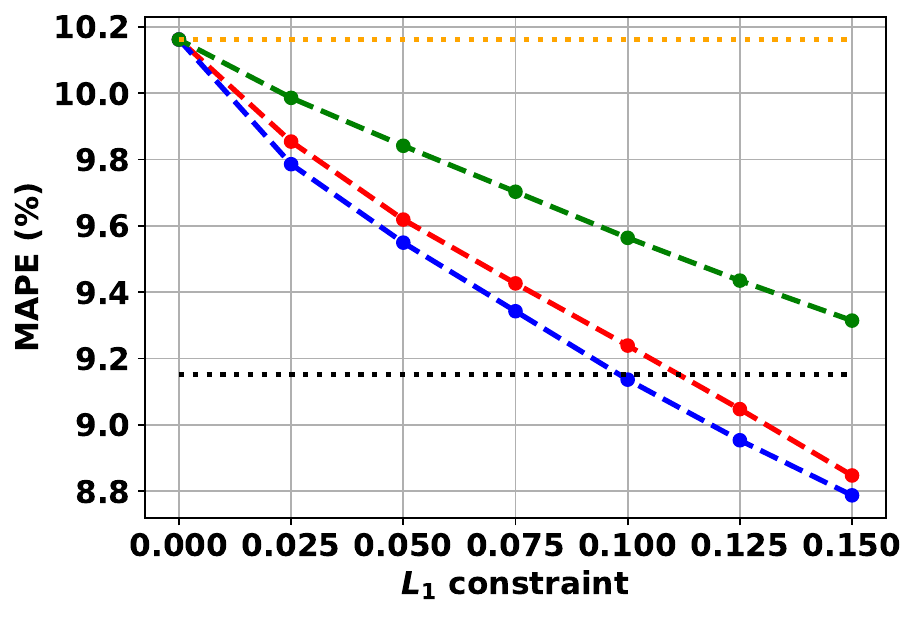}
         \label{fig:nn_conv_unchang_mape}
     \end{subfigure}
     \hfill
     \begin{subfigure}[b]{0.3\linewidth}
         \centering
         \includegraphics[width=\textwidth]{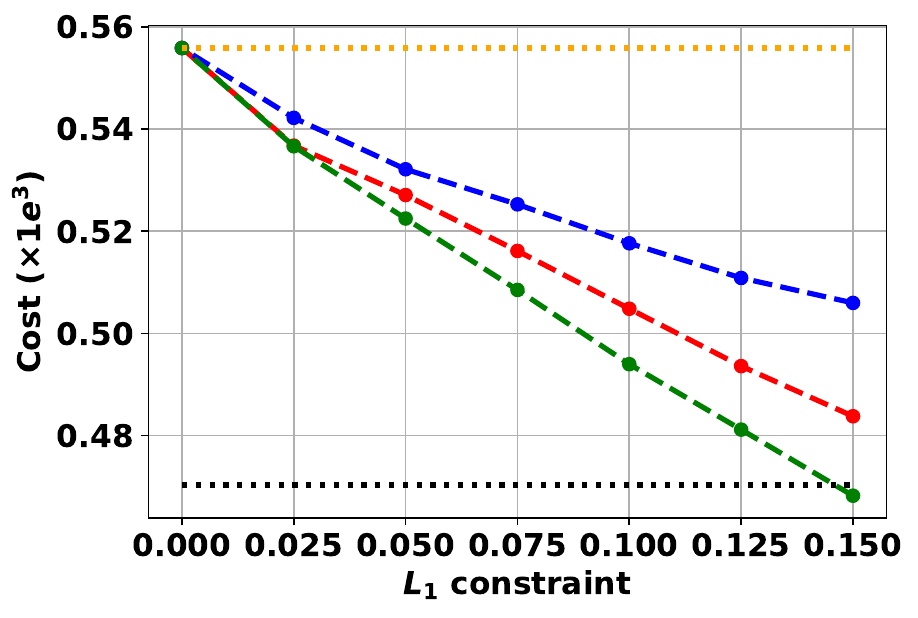}
         \label{fig:nn_conv_unchange_cost}
     \end{subfigure}
        \caption{Performance on the CNN load forecaster. First row: performance of complete machine unlearning algorithm \eqref{eq:unlearning}; Second row: Performance of PAMU and TAMU with different test criteria}
        \label{fig:nn_conv_unlearn_performance}
\end{figure*}

\begin{figure*}
     \centering
     \begin{subfigure}[b]{0.3\linewidth}
         \centering
         \includegraphics[width=\textwidth]{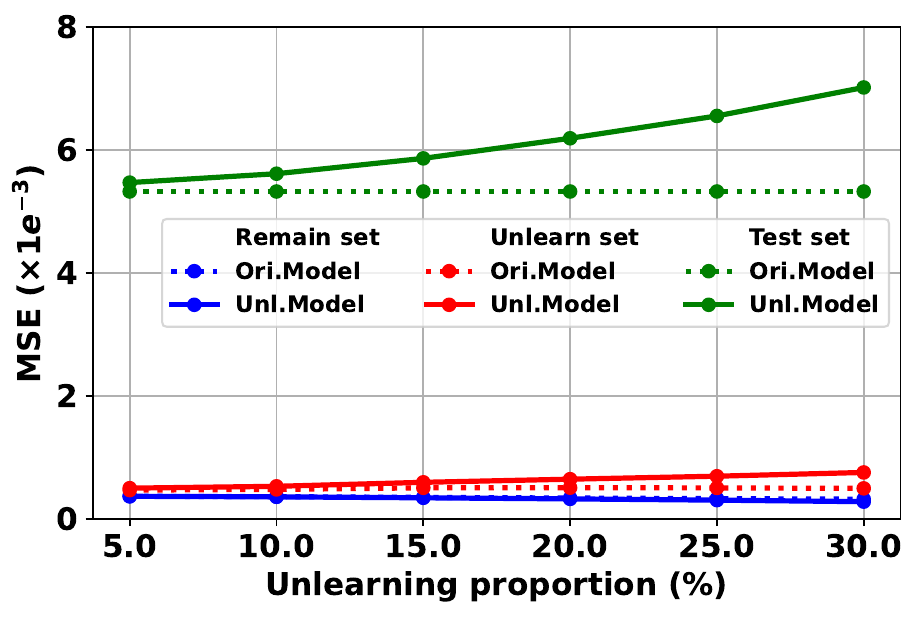}
         \label{fig:nn_mixer_unlearn_performance_mse}
     \end{subfigure}
     \hfill
     \begin{subfigure}[b]{0.3\linewidth}
         \centering
         \includegraphics[width=\textwidth]{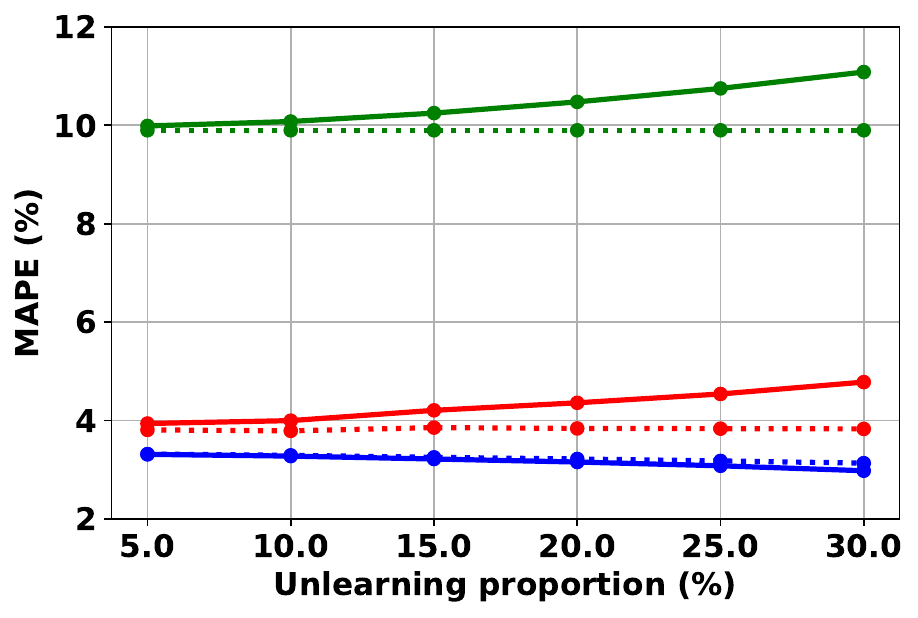}
         \label{fig:nn_mixer_unlearn_performance_mape}
     \end{subfigure}
     \hfill
     \begin{subfigure}[b]{0.3\linewidth}
         \centering
         \includegraphics[width=\textwidth]{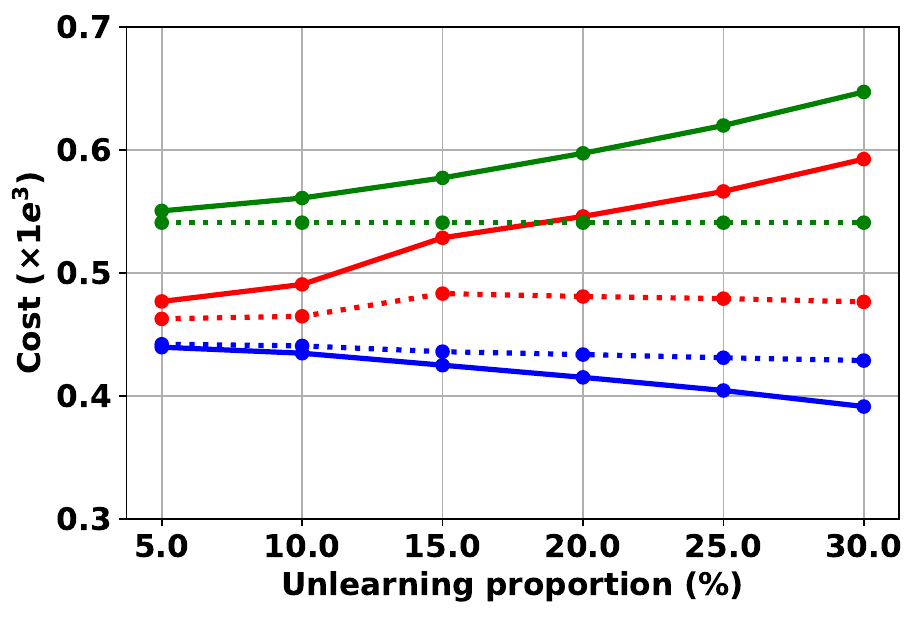}
         \label{fig:nn_mixer_unlearn_performance_cost}
     \end{subfigure}
     \hfill
     \begin{subfigure}[b]{0.3\linewidth}
         \centering
         \includegraphics[width=\textwidth]{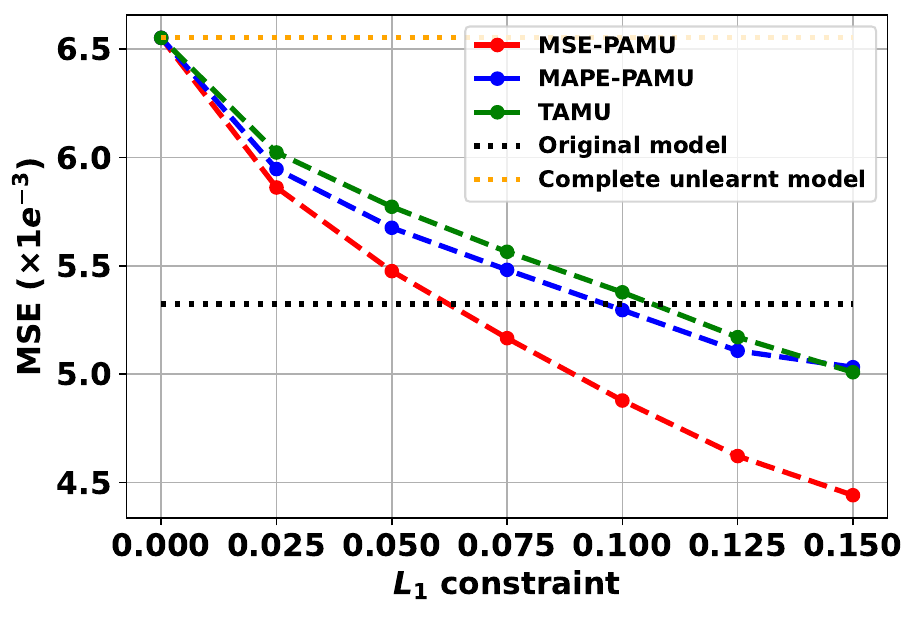}
         \label{fig:nn_mixer_unchange_mse}
     \end{subfigure}
     \hfill
     \begin{subfigure}[b]{0.3\linewidth}
         \centering
         \includegraphics[width=\textwidth]{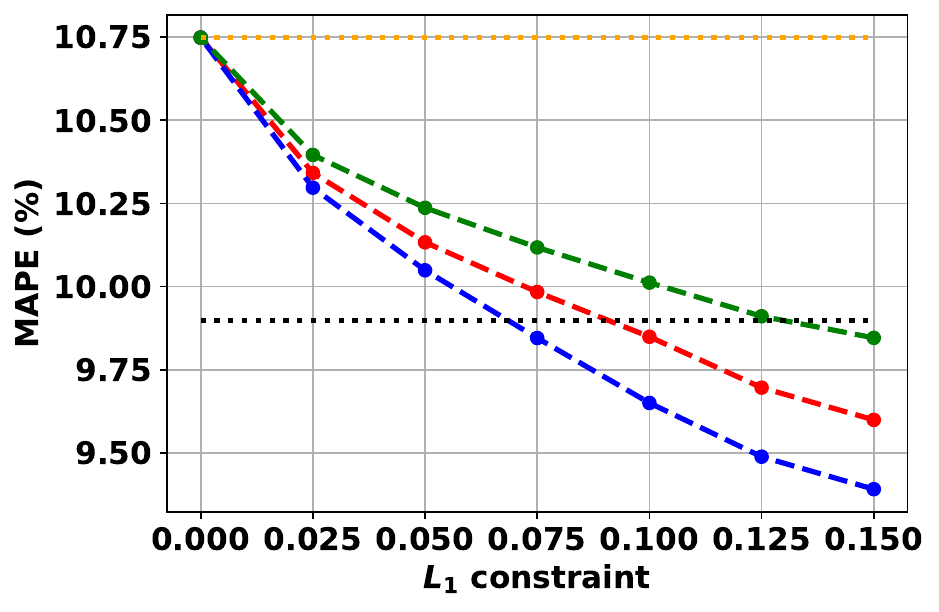}
         \label{fig:nn_mixer_unchang_mape}
     \end{subfigure}
     \hfill
     \begin{subfigure}[b]{0.3\linewidth}
         \centering
         \includegraphics[width=\textwidth]{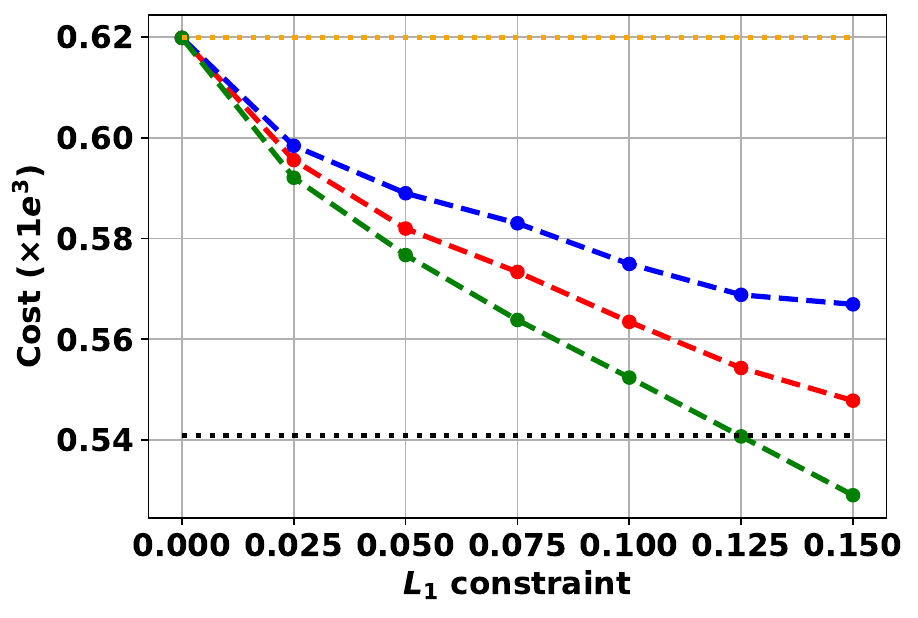}
         \label{fig:nn_mixer_unchange_cost}
     \end{subfigure}
        \caption{Performance on the MLP-Mixer load forecaster. First row: performance of complete machine unlearning algorithm \eqref{eq:unlearning}; Second row: Performance of PAMU and TAMU with different test criteria}
        \label{fig:nn_mixer_unlearn_performance}
\end{figure*}

The detailed unlearning performances on the CNN and MLP-Mixer based load forecasting models can be found in Fig.\ref{fig:nn_conv_unlearn_performance} and Fig.\ref{fig:nn_mixer_unlearn_performance}, respectively.

\newpage

\IEEEpeerreviewmaketitle

\bibliographystyle{IEEEtran}
\bibliography{IEEEabrv,Reference.bib}

\begin{thebibliography}{10}
\providecommand{\url}[1]{#1}
\csname url@samestyle\endcsname
\providecommand{\newblock}{\relax}
\providecommand{\bibinfo}[2]{#2}
\providecommand{\BIBentrySTDinterwordspacing}{\spaceskip=0pt\relax}
\providecommand{\BIBentryALTinterwordstretchfactor}{4}
\providecommand{\BIBentryALTinterwordspacing}{\spaceskip=\fontdimen2\font plus
\BIBentryALTinterwordstretchfactor\fontdimen3\font minus \fontdimen4\font\relax}
\providecommand{\BIBforeignlanguage}[2]{{%
\expandafter\ifx\csname l@#1\endcsname\relax
\typeout{** WARNING: IEEEtran.bst: No hyphenation pattern has been}%
\typeout{** loaded for the language `#1'. Using the pattern for}%
\typeout{** the default language instead.}%
\else
\language=\csname l@#1\endcsname
\fi
#2}}
\providecommand{\BIBdecl}{\relax}
\BIBdecl

\bibitem{xie2015long}
J.~Xie, T.~Hong, and J.~Stroud, ``Long-term retail energy forecasting with consideration of residential customer attrition,'' \emph{IEEE Transactions on Smart Grid}, vol.~6, no.~5, pp. 2245--2252, 2015.

\bibitem{hong2016probabilistic}
T.~Hong and S.~Fan, ``Probabilistic electric load forecasting: A tutorial review,'' \emph{International Journal of Forecasting}, vol.~32, no.~3, pp. 914--938, 2016.

\bibitem{hong2020energy}
T.~Hong, P.~Pinson, Y.~Wang, R.~Weron, D.~Yang, and H.~Zareipour, ``Energy forecasting: A review and outlook,'' \emph{IEEE Open Access Journal of Power and Energy}, vol.~7, pp. 376--388, 2020.

\bibitem{ebeid2017deducing}
E.~Ebeid, R.~Heick, and R.~H. Jacobsen, ``Deducing energy consumer behavior from smart meter data,'' \emph{Future Internet}, vol.~9, no.~3, p.~29, 2017.

\bibitem{luo2018benchmarking}
J.~Luo, T.~Hong, and S.-C. Fang, ``Benchmarking robustness of load forecasting models under data integrity attacks,'' \emph{International Journal of Forecasting}, vol.~34, no.~1, pp. 89--104, 2018.

\bibitem{liang2019poisoning}
Y.~Liang, D.~He, and D.~Chen, ``Poisoning attack on load forecasting,'' in \emph{2019 IEEE innovative smart grid technologies-Asia (ISGT Asia)}.\hskip 1em plus 0.5em minus 0.4em\relax IEEE, 2019, pp. 1230--1235.

\bibitem{wang2022personalized}
Y.~Wang, N.~Gao, and G.~Hug, ``Personalized federated learning for individual consumer load forecasting,'' \emph{CSEE Journal of Power and Energy Systems}, 2022.

\bibitem{dong2022short}
Y.~Dong, Y.~Chen, X.~Zhao, and X.~Huang, ``Short-term load forecasting with distributed long short-term memory,'' \emph{arXiv preprint arXiv:2208.01147}, 2022.

\bibitem{soykan2019differentially}
E.~U. Soykan, Z.~Bilgin, M.~A. Ersoy, and E.~Tomur, ``Differentially private deep learning for load forecasting on smart grid,'' in \emph{2019 IEEE Globecom Workshops (GC Wkshps)}.\hskip 1em plus 0.5em minus 0.4em\relax IEEE, 2019, pp. 1--6.

\bibitem{fernandez2022privacy}
J.~D. Fern{\'a}ndez, S.~P. Menci, C.~M. Lee, A.~Rieger, and G.~Fridgen, ``Privacy-preserving federated learning for residential short-term load forecasting,'' \emph{Applied Energy}, vol. 326, p. 119915, 2022.

\bibitem{husnoo2023secure}
M.~A. Husnoo, A.~Anwar, N.~Hosseinzadeh, S.~N. Islam, A.~N. Mahmood, and R.~Doss, ``A secure federated learning framework for residential short term load forecasting,'' \emph{IEEE Transactions on Smart Grid}, 2023.

\bibitem{mantelero2013eu}
A.~Mantelero, ``The eu proposal for a general data protection regulation and the roots of the ‘right to be forgotten’,'' \emph{Computer Law \& Security Review}, vol.~29, no.~3, pp. 229--235, 2013.

\bibitem{shaik2023exploring}
T.~Shaik, X.~Tao, H.~Xie, L.~Li, X.~Zhu, and Q.~Li, ``Exploring the landscape of machine unlearning: A survey and taxonomy,'' \emph{arXiv preprint arXiv:2305.06360}, 2023.

\bibitem{yu2023unlearning}
C.~Yu, S.~Jeoung, A.~Kasi, P.~Yu, and H.~Ji, ``Unlearning bias in language models by partitioning gradients,'' in \emph{Findings of the Association for Computational Linguistics: ACL 2023}, 2023, pp. 6032--6048.

\bibitem{zeng2023towards}
Y.~Zeng, J.~Xu, Y.~Li, C.~Chen, Q.~Dai, and Z.~Du, ``Towards highly-efficient and accurate services qos prediction via machine unlearning,'' \emph{IEEE Access}, 2023.

\bibitem{xia2023fedme}
H.~Xia, S.~Xu, J.~Pei, R.~Zhang, Z.~Yu, W.~Zou, L.~Wang, and C.~Liu, ``Fedme2: Memory evaluation \& erase promoting federated unlearning in dtmn,'' \emph{IEEE Journal on Selected Areas in Communications}, vol.~41, no.~11, pp. 3573--3588, 2023.

\bibitem{zhang2023poison}
Z.~Zhang, M.~Tian, C.~Li, Y.~Huang, and L.~Yang, ``Poison neural network-based mmwave beam selection and detoxification with machine unlearning,'' \emph{IEEE Transactions on Communications}, vol.~71, no.~2, pp. 877--892, 2023.

\bibitem{cao2015towards}
Y.~Cao and J.~Yang, ``Towards making systems forget with machine unlearning,'' in \emph{2015 IEEE Symposium on Security and Privacy}.\hskip 1em plus 0.5em minus 0.4em\relax IEEE, 2015, pp. 463--480.

\bibitem{ginart2019making}
A.~Ginart, M.~Guan, G.~Valiant, and J.~Y. Zou, ``Making ai forget you: Data deletion in machine learning,'' \emph{Advances in neural information processing systems}, vol.~32, 2019.

\bibitem{brophy2021machine}
J.~Brophy and D.~Lowd, ``Machine unlearning for random forests,'' in \emph{International Conference on Machine Learning}.\hskip 1em plus 0.5em minus 0.4em\relax PMLR, 2021, pp. 1092--1104.

\bibitem{golatkar2020eternal}
A.~Golatkar, A.~Achille, and S.~Soatto, ``Eternal sunshine of the spotless net: Selective forgetting in deep networks,'' in \emph{Proceedings of the IEEE/CVF Conference on Computer Vision and Pattern Recognition}, 2020, pp. 9304--9312.

\bibitem{peste2021ssse}
A.~Peste, D.~Alistarh, and C.~H. Lampert, ``Ssse: Efficiently erasing samples from trained machine learning models,'' \emph{arXiv preprint arXiv:2107.03860}, 2021.

\bibitem{fu2021bayesian}
S.~Fu, F.~He, Y.~Xu, and D.~Tao, ``Bayesian inference forgetting,'' \emph{arXiv preprint arXiv:2101.06417}, 2021.

\bibitem{golatkar2021mixed}
A.~Golatkar, A.~Achille, A.~Ravichandran, M.~Polito, and S.~Soatto, ``Mixed-privacy forgetting in deep networks,'' in \emph{Proceedings of the IEEE/CVF Conference on Computer Vision and Pattern Recognition}, 2021, pp. 792--801.

\bibitem{guo2019certified}
C.~Guo, T.~Goldstein, A.~Hannun, and L.~Van Der~Maaten, ``Certified data removal from machine learning models,'' \emph{arXiv preprint arXiv:1911.03030}, 2019.

\bibitem{bae2022if}
J.~Bae, N.~Ng, A.~Lo, M.~Ghassemi, and R.~B. Grosse, ``If influence functions are the answer, then what is the question?'' \emph{Advances in Neural Information Processing Systems}, vol.~35, pp. 17\,953--17\,967, 2022.

\bibitem{tuan2024learn}
T.~Hoang, S.~Rana, S.~Gupta, and S.~Venkatesh, ``Learn to unlearn for deep neural networks: Minimizing unlearning interference with gradient projection,'' in \emph{WACV}, Jan 2024.

\bibitem{graves2021amnesiac}
L.~Graves, V.~Nagisetty, and V.~Ganesh, ``Amnesiac machine learning,'' in \emph{Proceedings of the AAAI Conference on Artificial Intelligence}, vol.~35, no.~13, 2021, pp. 11\,516--11\,524.

\bibitem{bourtoule2021machine}
L.~Bourtoule, V.~Chandrasekaran, C.~A. Choquette-Choo, H.~Jia, A.~Travers, B.~Zhang, D.~Lie, and N.~Papernot, ``Machine unlearning,'' in \emph{2021 IEEE Symposium on Security and Privacy (SP)}.\hskip 1em plus 0.5em minus 0.4em\relax IEEE, 2021, pp. 141--159.

\bibitem{nguyen2022survey}
T.~T. Nguyen, T.~T. Huynh, P.~L. Nguyen, A.~W.-C. Liew, H.~Yin, and Q.~V.~H. Nguyen, ``A survey of machine unlearning,'' \emph{arXiv preprint arXiv:2209.02299}, 2022.

\bibitem{vohra2023end}
R.~Vohra, A.~Rajaei, and J.~L. Cremer, ``End-to-end learning with multiple modalities for system-optimised renewables nowcasting,'' \emph{arXiv preprint arXiv:2304.07151}, 2023.

\bibitem{xu2023e2e}
W.~Xu, J.~Wang, and F.~Teng, ``E2e-at: A unified framework for tackling uncertainty in task-aware end-to-end learning,'' \emph{arXiv preprint arXiv:2312.10587, acceped by AAAI-24}, 2023.

\bibitem{donti2017task}
P.~Donti, B.~Amos, and J.~Z. Kolter, ``Task-based end-to-end model learning in stochastic optimization,'' \emph{Advances in neural information processing systems}, vol.~30, 2017.

\bibitem{cook1982residuals}
R.~D. Cook and S.~Weisberg, \emph{Residuals and influence in regression}.\hskip 1em plus 0.5em minus 0.4em\relax New York: Chapman and Hall, 1982.

\bibitem{wu2022puma}
G.~Wu, M.~Hashemi, and C.~Srinivasa, ``Puma: Performance unchanged model augmentation for training data removal,'' in \emph{Proceedings of the AAAI Conference on Artificial Intelligence}, vol.~36, no.~8, 2022, pp. 8675--8682.

\bibitem{koh2017understanding}
P.~W. Koh and P.~Liang, ``Understanding black-box predictions via influence functions,'' in \emph{International conference on machine learning}.\hskip 1em plus 0.5em minus 0.4em\relax PMLR, 2017, pp. 1885--1894.

\bibitem{wang2023improving}
C.~Wang, Y.~Zhou, Q.~Wen, and Y.~Wang, ``Improving load forecasting performance via sample reweighting,'' \emph{IEEE Transactions on Smart Grid}, vol.~14, no.~4, pp. 3317--3320, 2023.

\bibitem{agrawal2019differentiable}
A.~Agrawal, B.~Amos, S.~Barratt, S.~Boyd, S.~Diamond, and J.~Z. Kolter, ``Differentiable convex optimization layers,'' \emph{Advances in neural information processing systems}, vol.~32, 2019.

\bibitem{zhuang2020comprehensive}
F.~Zhuang, Z.~Qi, K.~Duan, D.~Xi, Y.~Zhu, H.~Zhu, H.~Xiong, and Q.~He, ``A comprehensive survey on transfer learning,'' \emph{Proceedings of the IEEE}, vol. 109, no.~1, pp. 43--76, 2020.

\bibitem{pearlmutter1994fast}
B.~A. Pearlmutter, ``Fast exact multiplication by the hessian,'' \emph{Neural computation}, vol.~6, no.~1, pp. 147--160, 1994.

\bibitem{diamond2016cvxpy}
S.~Diamond and S.~Boyd, ``Cvxpy: A python-embedded modeling language for convex optimization,'' \emph{The Journal of Machine Learning Research}, vol.~17, no.~1, pp. 2909--2913, 2016.

\bibitem{lu2023synthetic}
J.~Lu, X.~Li, H.~Li, T.~Chegini, C.~Gamarra, Y.~Yang, M.~Cook, and G.~Dillingham, ``A synthetic texas backbone power system with climate-dependent spatio-temporal correlated profiles,'' \emph{arXiv preprint arXiv:2302.13231}, 2023.

\bibitem{tolstikhin2021mlp}
I.~O. Tolstikhin, N.~Houlsby, A.~Kolesnikov, L.~Beyer, X.~Zhai, T.~Unterthiner, J.~Yung, A.~Steiner, D.~Keysers, J.~Uszkoreit \emph{et~al.}, ``Mlp-mixer: An all-mlp architecture for vision,'' \emph{Advances in neural information processing systems}, vol.~34, pp. 24\,261--24\,272, 2021.

\bibitem{conejo2018power}
A.~J. Conejo and L.~Baringo, \emph{Power system operations}.\hskip 1em plus 0.5em minus 0.4em\relax Springer, 2018, vol.~11.

\bibitem{zhang2022cost}
J.~Zhang, Y.~Wang, and G.~Hug, ``Cost-oriented load forecasting,'' \emph{Electric Power Systems Research}, vol. 205, p. 107723, 2022.

\bibitem{jorge2006numerical}
N.~Jorge and J.~W. Stephen, \emph{Numerical optimization}.\hskip 1em plus 0.5em minus 0.4em\relax Spinger, 2006.

\end{thebibliography}

\bstctlcite{IEEEexample:BSTcontrol}


\begin{IEEEbiography}[{\includegraphics[width=1in,height=1.25in,clip,keepaspectratio]{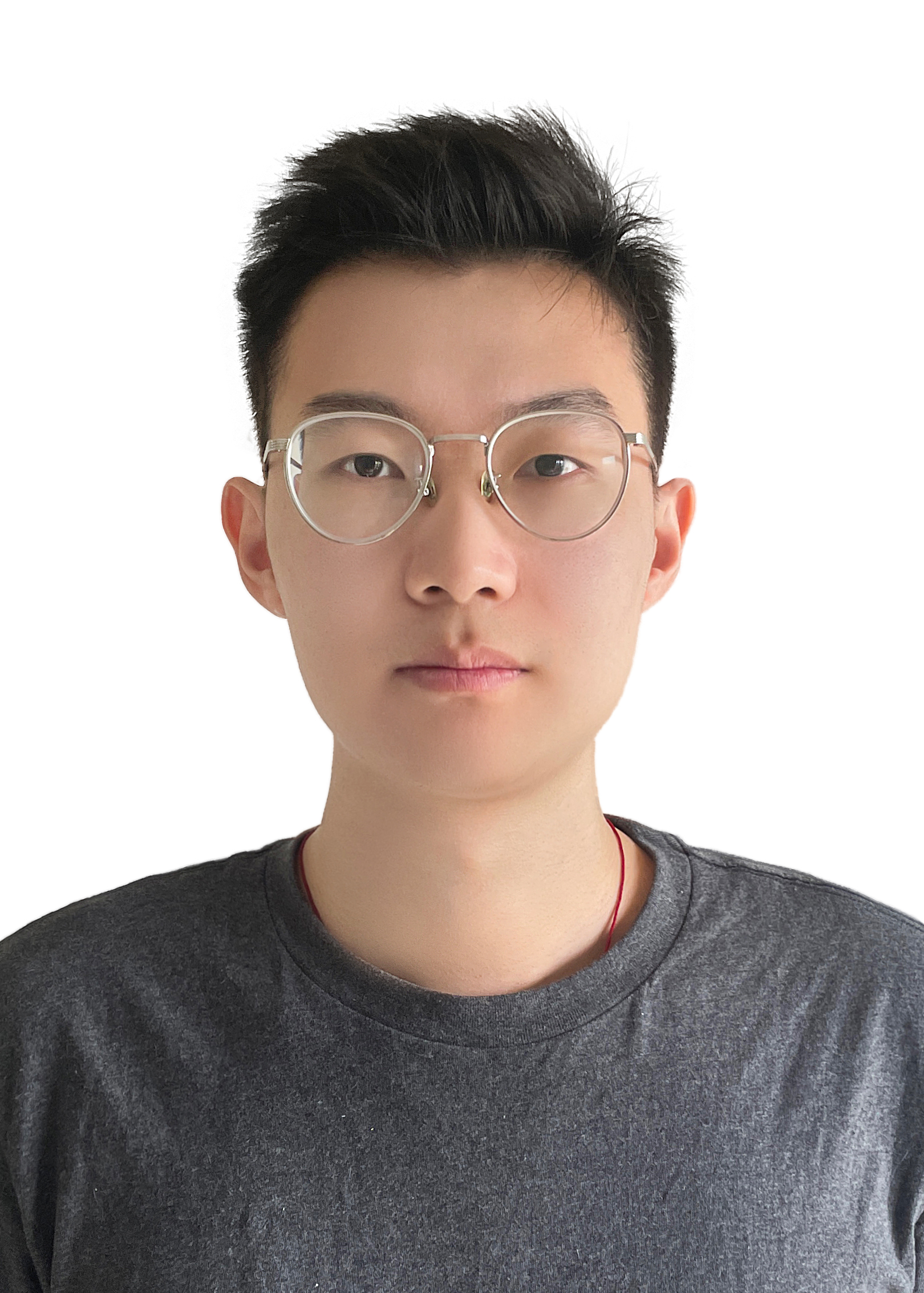}}]{Wangkun Xu} (Student Member, IEEE) received B.Eng. degree in electrical and electronic engineering from University of Liverpool, UK in 2018 and M.Sc. degree in control systems from Imperial College London, UK in 2019, where he is currently a Ph.D. student. His research focuses on machine learning and optimization, with application in cyber-physical power system operation and security.
\end{IEEEbiography}

\begin{IEEEbiography}[{\includegraphics[width=1in,height=1.25in,clip,keepaspectratio]{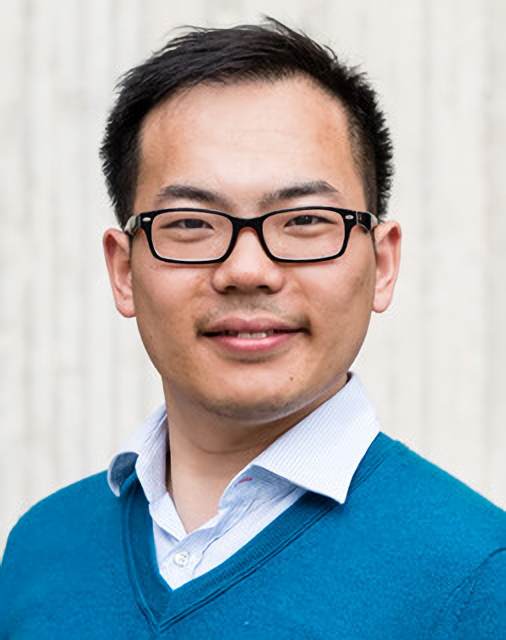}}]{Fei Teng} (Senior Member, IEEE) received the B.Eng. degree in electrical engineering from Beihang University, China, in 2009, and the M.Sc. and Ph.D. degrees in electrical engineering from Imperial College London, U.K., in 2010 and 2015, respectively, where he is currently a Senior Lecturer with the Department of Electrical and Electronic Engineering. His research focuses on the power system operation with high penetration of Inverter-Based Resources (IBRs) and the Cyber-resilient and Privacy-preserving cyber-physical power grid.
\end{IEEEbiography}

\ifCLASSOPTIONcaptionsoff
  \newpage
\fi
\end{document}